\documentclass{article}

\title{Multi-Task Interactive Robot Fleet Learning\\ with Visual World Models}

\usepackage[final]{corl_2024} 

\author{
  Huihan Liu
  \qquad 
  Yu Zhang
  \qquad
  Vaarij Betala
  \qquad
  Evan Zhang\\[5pt]
  \textbf{James Liu}
  \qquad
  \textbf{Crystal Ding}
  \qquad \textbf{Yuke Zhu} \\[5pt]
  The University of Texas at Austin
  \vspace{-4mm}
}

\usepackage{graphicx}
\usepackage{tikz}
\usepackage{float}
\usepackage{times}
\usepackage{latexsym}
\usepackage{amsmath}
\usepackage{amssymb}
\usepackage{amsfonts}
\usepackage{booktabs}
\usepackage{bbm}
\usepackage{xspace}
\usepackage{xcolor}

\usepackage{tabularx}
\usepackage{array}
\usepackage{caption}
\usepackage{longtable}
\usepackage{hyperref}

\newcolumntype{C}[1]{>{\centering\arraybackslash}p{#1}}

\newcommand{\methodname}[1]{\textsc{Sirius-Fleet}\xspace}

\usepackage{wrapfig}
\usepackage{subcaption}

\usepackage{listings}
\usepackage{siunitx}

\definecolor{mygreen}{rgb}{0,0.6,0}
\definecolor{mygray}{rgb}{0.5,0.5,0.5}
\definecolor{mymauve}{rgb}{0.58,0,0.82}
\definecolor{cerulean}{rgb}{0.0, 0.48, 0.65}
\definecolor{mycitecolor}{HTML}{007FFF}  

\usepackage{hyperref}       
\hypersetup{
    colorlinks,
    citecolor=mycitecolor,
    filecolor=black,
    linkcolor=mycitecolor,
    urlcolor=mycitecolor
}

\begin{document}
\maketitle


\begin{abstract}
    Recent advancements in large-scale multi-task robot learning offer the potential for deploying robot fleets in household and industrial settings, enabling them to perform diverse tasks across various environments. However, AI-enabled robots often face challenges with generalization and robustness when exposed to real-world variability and uncertainty. We introduce \methodname{}, a multi-task interactive robot fleet learning framework to address these challenges. \methodname{} monitors robot performance during deployment and involves humans to correct the robot's actions when necessary. We employ a visual world model to predict the outcomes of future actions and build anomaly predictors to predict whether they will likely result in anomalies. As the robot autonomy improves, the anomaly predictors automatically adapt their prediction criteria, leading to fewer requests for human intervention and gradually reducing human workload over time. Evaluations on large-scale benchmarks demonstrate \methodname{}'s effectiveness in improving multi-task policy performance and monitoring accuracy. We demonstrate \methodname{}'s performance in both RoboCasa in simulation and Mutex in the real world, two diverse, large-scale multi-task benchmarks. More information is available on the project website: \url{https://ut-austin-rpl.github.io/sirius-fleet}

\end{abstract}



\section{Introduction}

In recent years, there have been significant advancements in developing robots capable of performing various tasks~\cite{brohan2022rt1, reed2022generalist, shah2023vint}. The rapid progress in generalist robots holds great potential for deploying robot fleets \cite{hoque2023fleet, wang2024robot, müller2024openbotfleet} in households and industrial environments where the robots operate under a generalist multi-task policy. Despite these research advances, robots face challenges with generalization and robustness when deployed in real-world environments, which are often diverse and unstructured. These challenges undermine the safety and reliability of robot systems and limit their applicability in practical scenarios.

To address these challenges, a series of works have been developed on interactive imitation learning (IIL) \cite{mandlekar2020humanintheloop, 2022correct_me, spencer2020learning, hgdagger2019, hoque2021lazydagger, hoque2021thriftydagger, liu2022robot} and interactive fleet learning (IFL) \cite{hoque2022fleetdagger, wang2024robot, müller2024openbotfleet}. Prior work on human-in-the-loop learning \cite{mandlekar2020humanintheloop, liu2022robot, li2022efficient, hgdagger2019} has proposed involving humans in real-time monitoring and correction to ensure trustworthy deployment. However, these methods often require continuous human supervision. To reduce the high human workload, runtime monitoring approaches \cite{yel_runtime, liu2023modelbased, ood_survey, Richter_OOD, hoque2021thriftydagger, momart_wong22a, dass2022pato} have been proposed. These approaches automatically monitor robot performance, identify anomalies, and request human control when needed. Methods like out-of-distribution (OOD) detection \cite{momart_wong22a, dass2022pato, ood_survey} and failure detection \cite{hoque2021thriftydagger} have been introduced to identify anomaly cases of robot execution. However, these methods have primarily been used in single-task settings, limiting their effectiveness for large-scale, multi-task fleet deployment. 

\begin{figure*}[h!]
    \centering
    \vspace{1mm}
    \includegraphics[width=1\linewidth]{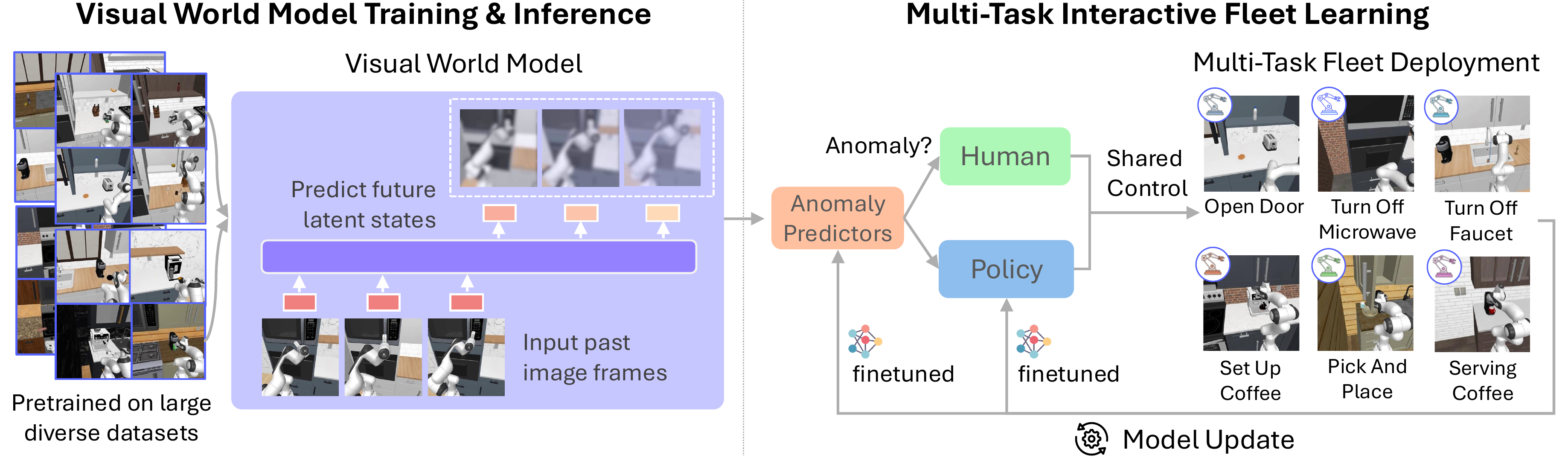}
    \caption{\textbf{Overview of \methodname{}.} Our framework of multi-task interactive robot fleet learning consists of two stages: 1) Visual World Model Training \& Inference, where we pre-train a visual world model on diverse datasets to predict future latent states from past visual observations, and 2) Multi-Task Interactive Fleet Learning, where anomaly predictors, built upon the pre-trained visual world model, enable real-time monitoring of the multi-task robot fleet during deployment, and solicit human feedback when necessary. The policy and anomaly predictors are continuously fine-tuned with deployment data, improving task performance over time.}
\label{fig:sample_figure}
\end{figure*}

We introduce {\methodname{}, a multi-task interactive robot fleet learning framework. \methodname{} consists of a multi-task policy and a runtime monitoring mechanism. \methodname{} enables the multi-task policy to be deployed across large robot fleets with the runtime monitoring mechanism, which requests human intervention when anomalies are predicted. \methodname{} tackles two challenges. First, \methodname{} enables a multi-task policy to be continually updated throughout long-term deployment. Multi-task policy training enables knowledge sharing across tasks and, therefore, improves generalization. Second, \methodname{} builds a runtime monitoring mechanism that efficiently supervises multiple tasks simultaneously. To address the first challenge, we train a multi-task policy that can be constantly fine-tuned with deployment data. For the second challenge, we employ a visual world model as the backbone for runtime monitoring, enabling the sharing of learned representations across downstream anomaly predictors for various tasks.

A key challenge in runtime monitoring is to effectively predict future action outcomes, as it allows the system to preempt potential failures before they occur. Recent advancements in world models \cite{videoworldsimulators2024, bruce2024genie, xiang2024pandora, hu2023gaia1} have shown the capability of simulating future scenarios and predicting future task outcomes. Inspired by this, we develop a visual world model trained on diverse robot trajectories performing a large variety of tasks, which enables the prediction of future task outcomes and helps prevent potential failures. The visual world model is trained by reconstructing image frames from input observations, which allows it to capture fine-grained visual details necessary for precise manipulation. The learned embeddings from the world model are then shared across downstream anomaly prediction tasks. We train two distinct types of anomaly predictors: failure prediction and out-of-distribution (OOD) prediction. The two predictors complement each other --- failure prediction predicts failures similar to those identified by humans previously, and OOD prediction captures cases when the robot is in novel, unfamiliar scenarios.
The predictors are trained using the frozen embeddings from the visual world model and are continuously fine-tuned during deployment. Unlike prior work that uses fixed prediction thresholds, \methodname{} automatically adjusts the anomaly predictors' criteria based on task performance and human feedback. This adaptive threshold aligns with the robot's evolving level of autonomy, resulting in more effective runtime monitoring.

We evaluate our multi-task interactive robot fleet learning framework on large-scale benchmarks in simulation and real-world environments. Our key findings are: 1) our runtime monitoring system effectively supervises diverse multi-task scenarios with $>95\%$ success rates in overall system performance, 2) the multi-task policy continually improves over time by leveraging deployment data, 3) our anomaly predictors for runtime monitoring outperform baseline methods in accuracy and reduction of human workload. In summary, our contributions are:

\begin{enumerate}
    \item A framework for multi-task interactive robot fleet learning. The multi-task robot policy efficiently improves over deployment through runtime monitoring and human interaction;
    \item A runtime monitoring mechanism based on a visual world model backbone with task-adaptive anomaly prediction thresholds;
    \item A demonstration of the high performance of our multi-task fleet learning system in both simulation and real-world environment, achieving on average $>95\%$ success rates in system performance.
\end{enumerate}

\section{Related Work}

\textbf{Multi-Task Robot Learning.} Recent advancements in multi-task robot learning have seen the development of agents capable of performing diverse tasks across various domains \cite{reed2022generalist, jiang2023vima, shah2023mutex, brohan2022rt1, brohan2023rt2, jang2022bcz, bousmalis2023robocat, octomodelteam2024octo} These advances have been driven by innovations in policy architectures \cite{brohan2022rt1, robomimic, ho2020denoising, chi2024diffusion}, the availability of large-scale datasets \cite{ebert2021bridge, open_x_embodiment_rt_x_2023, khazatsky2024droid, mandlekar2023mimicgen}, and new robotics benchmarks \cite{robomimic, robocasa2024}. Despite this progress, many of these models are deployed as static, one-off implementations, which limits their robustness and generalization in real-world, unstructured environments. In contrast, \methodname{} is the first framework for robot manipulation that enables iterative improvement of multi-task policies through human-in-the-loop interaction during deployment.

\textbf{Robot Fleet Learning.} The recent large-scale robot deployment and data collection efforts \cite{kalashnikov2018qt, aloha2team2024aloha, ahn2024autort} have increased interest in robot fleet learning \cite{swamy2020scaled, hoque2023fleet, wang2024robot, akcin23a_decentral, akcin23a_fleet_active, müller2024openbotfleet}. This area of research addresses key challenges such as resource allocation \cite{swamy2020scaled, hoque2022fleetdagger}, decentralized and federated learning \cite{wang2024robot, akcin23a_decentral, akcin23a_fleet_active}, and system management \cite{müller2024openbotfleet}. However, two significant challenges remain: supervising large robot fleets with minimal human oversight and enabling multi-task policy improvement over time using deployment data. \methodname{} addresses these challenges by combining runtime monitoring with continuous policy updates in a multi-task, interactive fleet learning framework. This ensures that policies improve iteratively based on deployment data and human feedback.

\textbf{Interactive Imitation Learning.} Human-in-the-loop methods have been introduced to ensure safe and trustworthy deployment by allowing robots to learn from human interventions during task execution \cite{hgdagger2019, spencer2020learning, mandlekar2020humanintheloop, liu2022robot, li2022efficient, 2022correct_me, hoque2021lazydagger}. To reduce the burden of constant human oversight, runtime monitoring techniques have been developed to identify anomalies during task execution \cite{hoque2021thriftydagger, liu2023modelbased, Hsu2023TheSF, yel_runtime, SinhaSchmerlingEtAl2023, Nakamura2024AGC}. Two main areas of focus are unsupervised out-of-distribution (OOD) detection \cite{ood_survey, sinha2022systemlevel, Richter_OOD, menda2019ensembledagger, momart_wong22a, dass2022pato}, and failure detection, which can be done either by binary classification \cite{diryag2014neural,  Gokmen2023AskingFH, xie2022ask} or by learning risk functions from trajectory data ~\cite{hoque2021thriftydagger, hoque2022fleetdagger}. Recent advancements in foundation models have also led to the use of Large Language Models (LLMs) and Vision Language Models (VLMs) to identify anomalies for policies based on these models \cite{ren2023robots, liu2023reflect, guan2024task}. While prior work has primarily focused on task-specific dynamics models for single-task settings, \methodname{} introduces a visual world model trained on diverse datasets before deployment. This model can predict anomalies across various tasks, making \methodname{} scalable without additional training during deployment.
\section{\methodname{}: Multi-Task Interactive Robot Fleet Learning}

\subsection{Background}

\subsubsection{Problem Formulation}

We formulate multi-task interactive robot fleet learning as a finite-horizon Markov Decision Process (MDP), where $N$ robots operate in $N$ independent MDPs. The $i$-th robot operates in its respective $i$-th MDP, defined as $\mathcal{M}_i = (S, A, \mathcal{T}, H_i, \mu^0_i, R_i)$, where $S$ is the state space, $A$ is the action space, $\mathcal{T} : S \times A \rightarrow S$ is the transition dynamics, $H_i$ is the horizon length, $\mu^0_i$ is the initial state distribution, and  $R_i : S \times A \rightarrow \mathbb{R}$ is the reward function. In sparse-reward settings, $R_i$ is replaced with a goal predicate $g_i : S \rightarrow \{0, 1\}$. The collection of MDPs, $\{\mathcal{M}_i\}^N_{i=1}$, can be reformulated as a single unified MDP with shared state space $S$, action space $A$, and transition function $\mathcal{T}$. Data from all robots are aggregated to learn a unified multi-task policy $\pi(a \mid s, g_i)$, which maximizes the expected return: $\max_{\pi} J(\pi) = \mathbb{E}_{s_t, a_t \sim \pi, \mu_0} \left[\sum_{t=1}^{H} g_i(s_t)\right]$. 

\subsubsection{Multi-Task Interactive Fleet Deployment}
\label{deploy-workflow}

We consider an interactive learning framework \cite{mandlekar2020humanintheloop, liu2022robot, liu2023modelbased} for a fleet of robots \cite{hoque2023fleet}, where learning and deployment happens iteratively with a human policy $\pi_H$ in the loop. Each robot bootstraps its initial policy $\pi_0$ via behavioral cloning from a set of human demonstrations, $\mathcal{D}^0$. Starting from deployment round $i=1$, each robot executes tasks using policy $\pi_i$ with runtime monitoring, supervised by anomaly predictors $E_i$. 
During policy execution, $E_i$ determines whether the current state may lead to anomalies. Upon predicting a potential anomaly,  the system signals the human to monitor the process. While monitoring, the human can choose to actively intervene \cite{hu2022modelbased} and take control if necessary, allowing the human policy $\pi_H(s,g)$ to override the robot policy $\pi_i(s,g)$. The resulting trajectories $\tau = (s_t, a_t, r_t, c_t)$ are added to the data buffer $\mathcal{D}'$ of the current deployment round, where $c_t$ indicates whether timestep $t$ is controlled by human action. The data buffer is then updated as $\mathcal{D}^{i+1} = \mathcal{D}^{i} \cup \mathcal{D}'$, which is used to train the next-round policy $\pi_{i+1}$ and anomaly predictors $E_{i+1}$.

\begin{figure*}[t]
    \centering
    \includegraphics[width=1\linewidth]{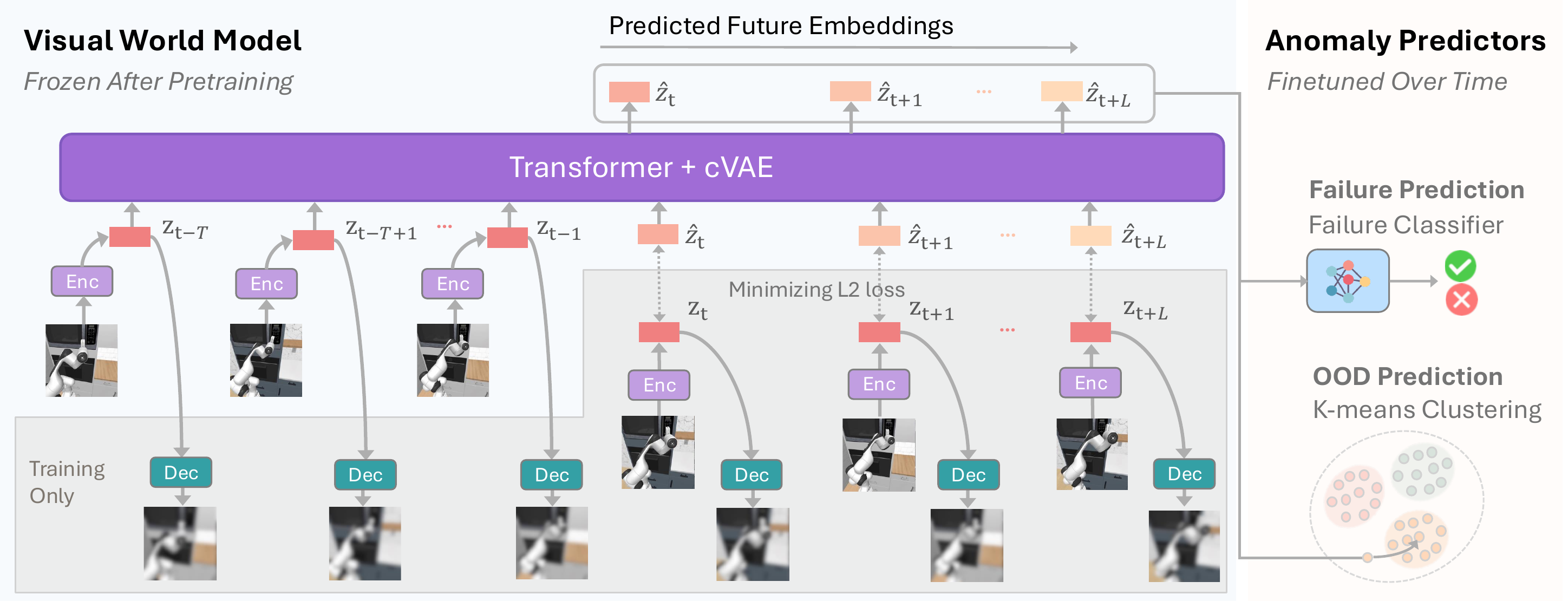}
    \caption{\textbf{Model Architecture.} The visual world model comprises a UNet-based encoder and decoder combined with a cVAE- and Transformer-based prediction model. This architecture allows the world model to predict future embeddings from the current state. The learned representations are then used for anomaly predictions, including failure and OOD prediction.}
    \vspace{-4mm}
    \label{fig:model_architecture}
\end{figure*}

\subsection{Runtime Monitoring for Multi-Task Fleet Deployment}

We present \methodname{}'s runtime monitoring mechanism, which supervises multiple tasks simultaneously across diverse environments during deployment. The system is designed to meet three key goals: 1) generalization---the anomaly predictors are built on shared embeddings from the visual world model, allowing them to be used across different tasks; 2) task adaptability---it adjusts dynamically to the evolving progress of each task during deployment; and 3) failure preemption---it predicts anomalies before they occur, enabling timely prediction and intervention. To achieve this, we train a visual world model that simulates task progress and supports anomaly prediction across various tasks, as illustrated in Figure \ref{fig:model_architecture}.

\subsubsection{Training the Visual World Model}

Inspired by recent advances in world models \cite{videoworldsimulators2024, bruce2024genie, xiang2024pandora, hu2023gaia1}, we train a visual world model on diverse robot trajectory frames to predict future task outcomes and prevent potential failures. The visual world model, trained by reconstructing image observations, predicts future latent embeddings. This world modeling approach is effective for several reasons: 1) pixel reconstruction is a readily available form of supervision, allowing the model to learn without manual annotations; 2) reconstructing image frames helps the model capture the fine-grained visual details necessary for precise manipulation tasks, and 3) it helps to develop the model's ability to predict changes in the robot's visual environment over time.

\methodname{} trains an autoregressive visual world model $\mathcal{W}$ as the backbone for downstream anomaly prediction. The world model $\mathcal{W} = (E_\gamma, D_\lambda, T_\psi)$ consists of an encoder $E_\gamma$, a decoder $D_\lambda$, and a conditional next state prediction model $T_\psi$. In training, $E_\gamma$ encodes image observation $x_t$ at timestep $t$ into latent embedding $z_t$. The decoder $D_\lambda$ reconstructs $z_t$ to image $\hat{x_t}$ and minimizes the image reconstruction L2 loss. $T_\psi$ inputs the history of $T$ timesteps of embeddings $(z_{t-T}, z_{t-T+1}, \dots z_t)$, and outputs $\hat{z_{t+1}}$. It autoregressively predicts $(\hat{z_{t+1}}, \hat{z_{t+2}}, \dots \hat{z_{t+L}})$ for $L$ steps into the future using the last $T$ timesteps of embeddings and reconstructed embeddings, and minimizes the embedding reconstruction loss between $(\hat{z_{t+1}}, \hat{z_{t+2}}, \dots \hat{z_{t+L}})$ and $(z_{t+1}, z_{t+2}, \dots z_{t+L})$. $E_\gamma$ and $D_\lambda$ are implemented with UNet \cite{ronneberger2015unet, ho2020denoising}, and  $T_\psi$ uses conditional Variational Autoencoder (cVAE) \cite{NIPS2015_8d55a249}. $T_\psi$ is jointly trained with $E_\gamma, D_\lambda$ on the same latent space. 

We use a stochastic latent space rather than a deterministic one since the stochastic latent space of cVAE supports multiple future sampling and facilitates better prediction \cite{wu2023daydreamer, liu2023modelbased}. We use transformer architecture \cite{vaswani2023attention} for the encoder, decoder, and prior network for the cVAE in $T_\psi$.

\subsubsection{Building the Downstream Anomaly Predictors}
\vspace{-2mm}
The visual world model captures changes in task outcomes over time. Its learned representation can be used for the anomaly predictors to predict future anomalies. Since individual tasks vary, we train task-specific anomaly predictors on the frozen representation to predict failures and OOD anomalies.

\begin{wrapfigure}[25]{r}{0.5\textwidth}  
    \centering
    \vspace{-1mm}
    \begin{subfigure}[b]{\linewidth}
        \centering
        \includegraphics[width=\linewidth]{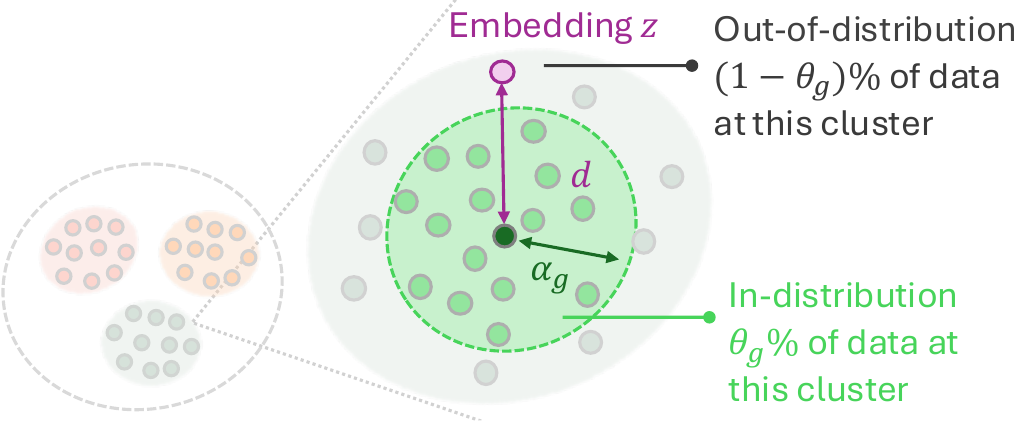} 
    \end{subfigure}
    
    \vspace{-0.6mm} 
    
    \begin{subfigure}[b]{\linewidth}
        \centering
        \includegraphics[width=\linewidth]{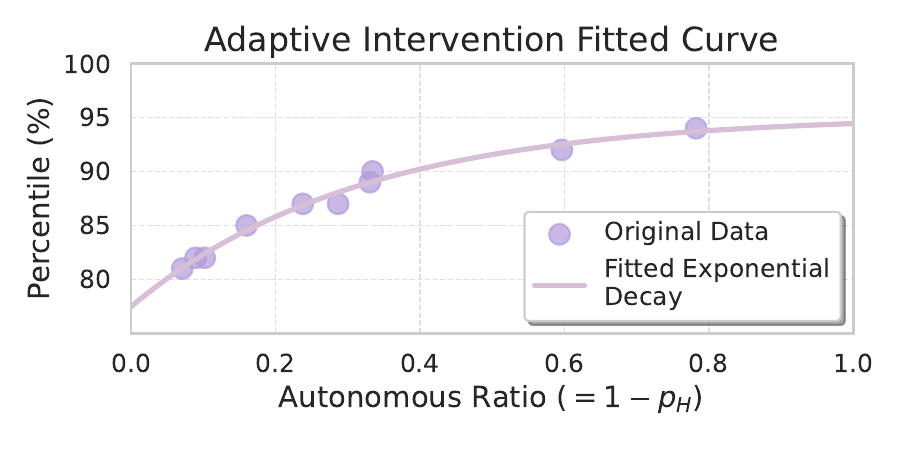} 
    \end{subfigure}
    
    \caption{\textbf{Adaptive Decision Boundaries.} Top: OOD Prediction Boundary. The threshold $\theta_g$, determined by the human intervention ratio, sets the distance threshold $\alpha_g$. A sample is identified as OOD if its embedding's distance $d$ from the cluster centroid exceeds $\alpha_g$. Bottom: Fitting function for optimal $\theta_g$ based on the human intervention ratio $p_H$. The x-axis shows $1 - p_H$, representing the autonomous rollout ratio.}

    \label{fig:adaptive}
\end{wrapfigure}

\textbf{Failure Prediction.} We train a failure classifier $F_\sigma$ for each task using the frozen image embeddings from the visual world model. Failure labels are from trajectories $\tau = (s_t, a_t, r_t, c_t)$ in the last round $\mathcal{D}'$ (see Section \ref{deploy-workflow}), where $c_t$ marks human interventions (\texttt{human}). The trajectory segment before each intervention are labeled as failures (\texttt{failure}) \cite{liu2022robot, liu2023modelbased}. The classifier is trained using a cross-entropy loss $\mathcal{L}_{\text{{F}}} = -\sum_{i=1}^{n} y_i \log(\hat{y}_i)$ with balanced sampling, where $y_i \in \{\texttt{rollout}, \texttt{failure}, \texttt{human}\}$. The failure classifier is a small, computationally efficient model trained on frozen world model embeddings (training time $1.5$ hour).

\textbf{Out-of-Distribution (OOD) Prediction.} We identify OOD states using k-means clustering. The frozen visual world model $\mathcal{W}$ generates embeddings from sampled trajectories in the data buffer. We use Principal Component Analysis (PCA) to reduce the embedding dimensions to $l$ and calculate $c$ k-means centroids for each task. To predict OOD for an embedding $z$, we reduce its dimensions, find the nearest centroid, and calculate its L2 distance $d$. A state is identified as OOD if $d$ exceeds the task threshold $\alpha_g$.  $\alpha_g$ is determined by $d^\theta_g$, which is the distance of the top $\theta_g$ percentile of the distances to the nearest centroids from the validation latent embeddings. For efficiency, we perform task-specific k-means clustering on image embeddings for the specific task without training additional models. 

\textbf{Adaptive Decision Boundaries.} \label{method: adaptive} Anomaly predictors should adjust their anomaly prediction threshold as robot performance improves during deployment. In a multi-task setting, the varying task performances across different tasks pose additional challenges. We adjust anomaly predictors by loosening the decision boundaries for high-performing tasks and tightening them for low-performing tasks. For failure prediction, we finetune it using the most recent round of deployment data $\mathcal{D}'$, with human intervention labels reflecting the updated human perceived risk.

The decision boundary expands for OOD prediction as the robot encounters more in-distribution states. 
We sample from $D^i$, the aggregated deployment data for all rounds, for k-means clustering. For each new embedding $z$, we use a threshold $\alpha_g$ for its distance $d$ to its nearest centroid, which depends on $\theta_g$. We update the prediction threshold $\alpha_g$ based on the human intervention ratio $p_H$, which acts as a proxy for policy performance. We fit the exponential decay function, $\theta_g = a + b e^{c p_H}$, and obtain the parameters by fitting the curve to a set of calibration trajectories. Empirically, $a = 95.2$, $b = -17.7$ and $c = -3.2$. We found this function to be robust across all rounds and all tasks from simulation to real-world experiments, and we used the same function and set of hyperparameters throughout the simulation and real-world experiments. 

\begin{wrapfigure}[17]{r}{0.45\textwidth}
\vspace{-3mm}
    \centering
    \includegraphics[width=0.45\textwidth]{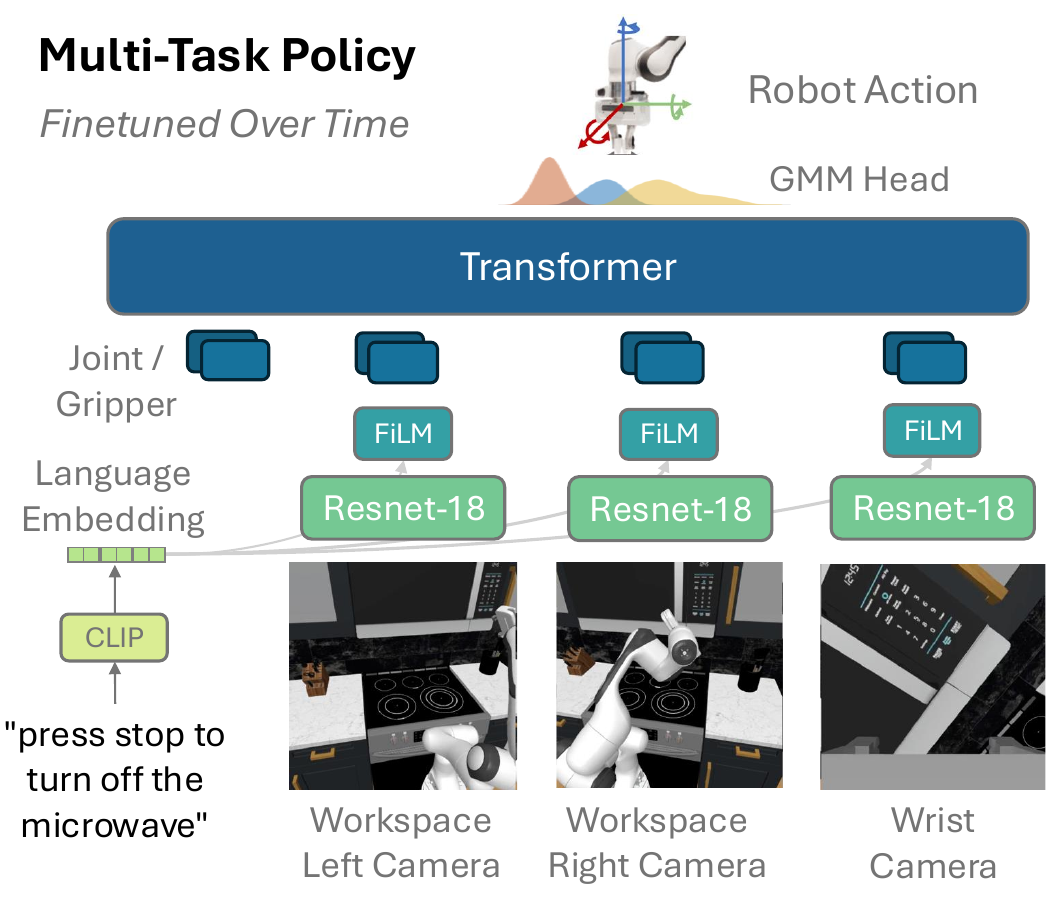}
    \caption{\textbf{Policy Architecture.} The multi-task policy is a Transformer that processes images, proprioceptive data, and task language embeddings. It uses a Gaussian Mixture Model (GMM) to output robot actions.}
    \label{fig:policy}
\end{wrapfigure}

At deployment, the embedding $z$ is computed from observations, and the distance $d$ from the nearest centroid is compared to $\alpha_g$. If $d > \alpha_g$, the sample is identified as OOD. Figure \ref{fig:adaptive} (Top) shows the decision boundary, while Figure \ref{fig:adaptive} (Bottom) shows the fitted curve for $\theta_g$. 

\subsubsection{\methodname{} in Operation}

\textbf{Continual Model Improvement.} While the visual world model $\mathcal{W}$ is trained once and frozen, the policy and the anomaly predictors are continually finetuned over deployment: the policy is updated with the data from previous rounds using weighted sampling, the failure classifier for failure prediction is finetuned with the most recent intervention labels, and the K-means clustering in OOD prediction expands its latent space coverage after each round. 

\textbf{Anomaly Predictors at Runtime.} During deployment, the visual world model predicts each task's future embeddings over $L$ steps. We generate $N$ possible future scenarios by sampling from the cVAE latent space and making $L$-step predictions $N$ times. Each predicted future embedding is then assessed individually by the anomaly predictors. For failure prediction, we compute the average failure score across future embeddings. For OOD prediction, we calculate the average distance of each future state to its nearest cluster centroid and compare the final average distance against the OOD threshold.

\textbf{Multi-Task Policy Training.} We train a Transformer-based \cite{vaswani2023attention} multi-task policy, as shown in Figure \ref{fig:policy}. The policy inputs image observations, robot proprioceptive data, language task goals, and output robot actions. The design follows that in RoboMimic \cite{robomimic} and RoboCasa \cite{robocasa2024}.

\section{Experiments}
\label{exp:main}

In our experiments, we aim to address the following questions: 
1) How well does \methodname{} continue to improve its policy during deployment with effective runtime monitoring? 2) How does \methodname{}'s runtime monitoring performance compare with baseline methods? 3) Are the anomaly prediction made by \methodname{} at the right moments, qualitatively?

\begin{figure*}[t]
    \centering
    \includegraphics[width=0.95\linewidth]{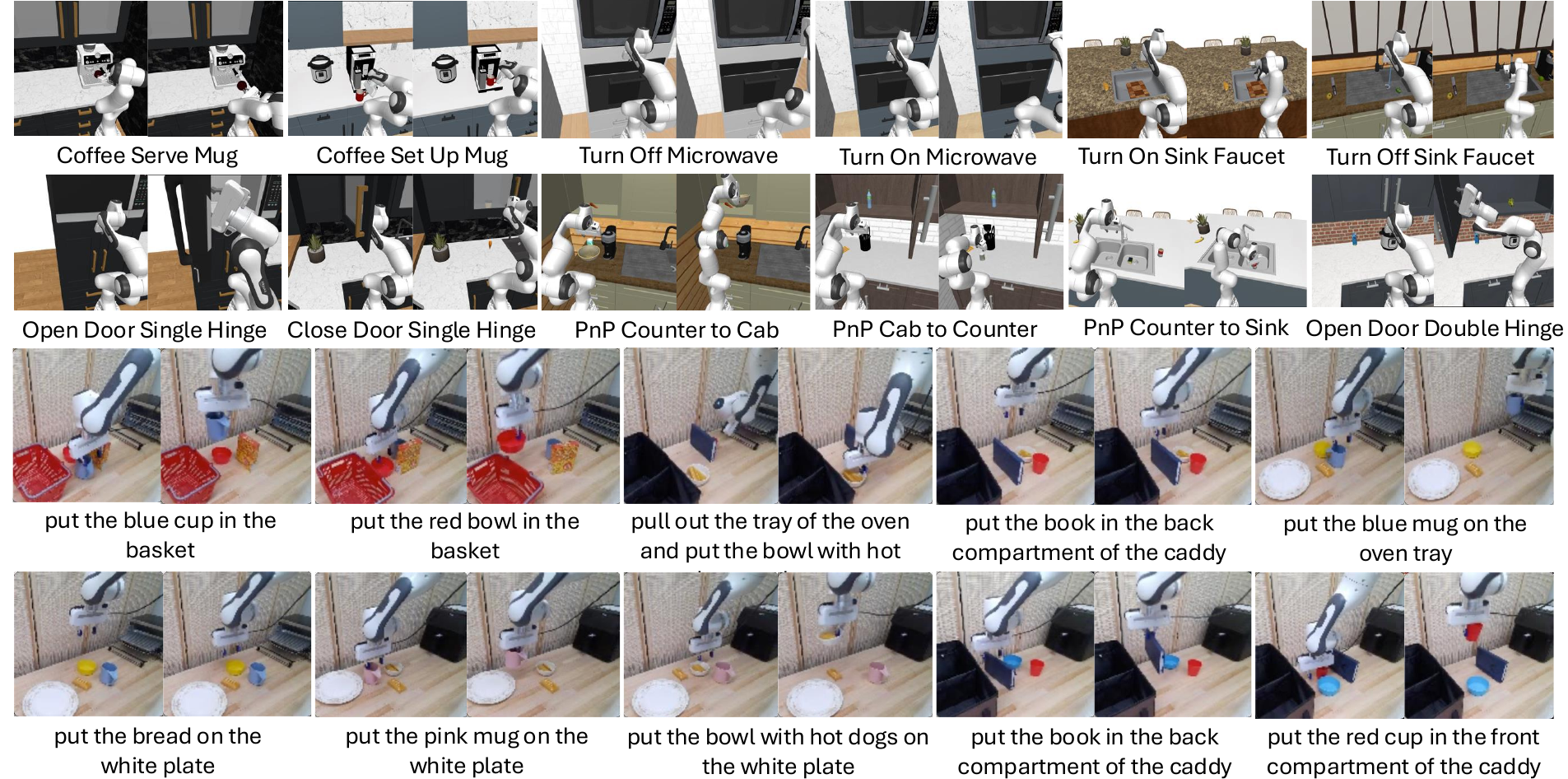}
    \caption{\textbf{RoboCasa Simulation Tasks and Mutex Real-World Tasks.} We evaluate policy learning and runtime monitoring using 12 tasks from the RoboCasa benchmark in simulation and 10 tasks from the Mutex benchmark in real-world environments.}
    \label{fig:task_figure}
\end{figure*}

\subsection{Evaluation Setup}

\label{eval_setup}

To measure the effectiveness of \methodname{}'s multi-task fleet learning, we evaluate how both the policy and the runtime monitoring perform and evolve over time. Following prior works \cite{mandlekar2018roboturk, liu2022robot, liu2023modelbased}, we evaluate the system in a human-in-the-loop setting through rounds of iterative deployment. 

\textbf{Evaluation Setting.} \label{exp:evaluation_setting}
Our evaluation spans three rounds of policy updates and runtime monitoring. Full human supervision is required in the first round since no anomaly predictors have been trained yet. Human supervision is only requested in the next two rounds when an anomaly is predicted. After each round, the collected deployment data and intervention labels are used to train the policy and anomaly predictors for the next round. The initial policy is trained with $50$ human demonstrations per task over 1200 epochs. We collect $100$ rollouts per task for each round to ensure consistent data size and finetune the policy for $400$ epochs.  

\textbf{Evaluation Metrics.} \label{exp:evaluation_metrics}
We evaluate \methodname{}'s policy using \textbf{Autonomous Performance}, which measures the policy's ability to achieve its goals without human intervention. After each round, the policy is finetuned on the newly collected data, and we evaluate its success rate without runtime monitoring to assess performance improvements over time.
To evaluate runtime monitoring, we use \textbf{Combined Policy Performance (CPP)} and \textbf{Return of Human Effort (ROHE)}, following \cite{hoque2021thriftydagger, liu2023modelbased}. CPP measures the overall system success rate under monitoring, reflecting the effectiveness of human-robot collaboration in preventing anomalies. ROHE assesses the efficiency of human intervention by comparing the policy's success rate against the amount of human intervention effort, calculated as the ratio of policy performance to the number of interventions: $\text{Normalized ROHE}=\frac{\mathbb{E}_\tau[\sum_{t=0}^{T_{\tau}}r^\tau_t]}{1+\frac{H}{T}}$. The goal is to maximize performance while minimizing human workload.

\textbf{Evaluation Environments.} \label{exp:evaluation_envs}
For simulation, we use RoboCasa \cite{robocasa2024}, a visually diverse benchmark with various objects, layouts, and scenes. RoboCasa contains both a large dataset using MimicGen \cite{mandlekar2023mimicgen}, an automated trajectory generation method (5k trajectories per task), and a human demonstration dataset (50 human demonstrations per task). The visual world model is trained on 20 task suites of MimicGen data, and the policy is trained on 12 task suites of human demonstration data. We use Mutex \cite{shah2023mutex} for real-world experiments, a commonly used benchmark for multi-task learning. The visual model is trained on 50 tasks, and the policy is trained on 10 tasks across 5 task suites. We use an OSC controller with 7D action space (x-y-z position, yaw-pitch-roll orientation, and gripper).

\begin{figure}[t]
    \centering
    \begin{subfigure}[t]{0.305\textwidth}
        \centering
        \includegraphics[width=\textwidth]{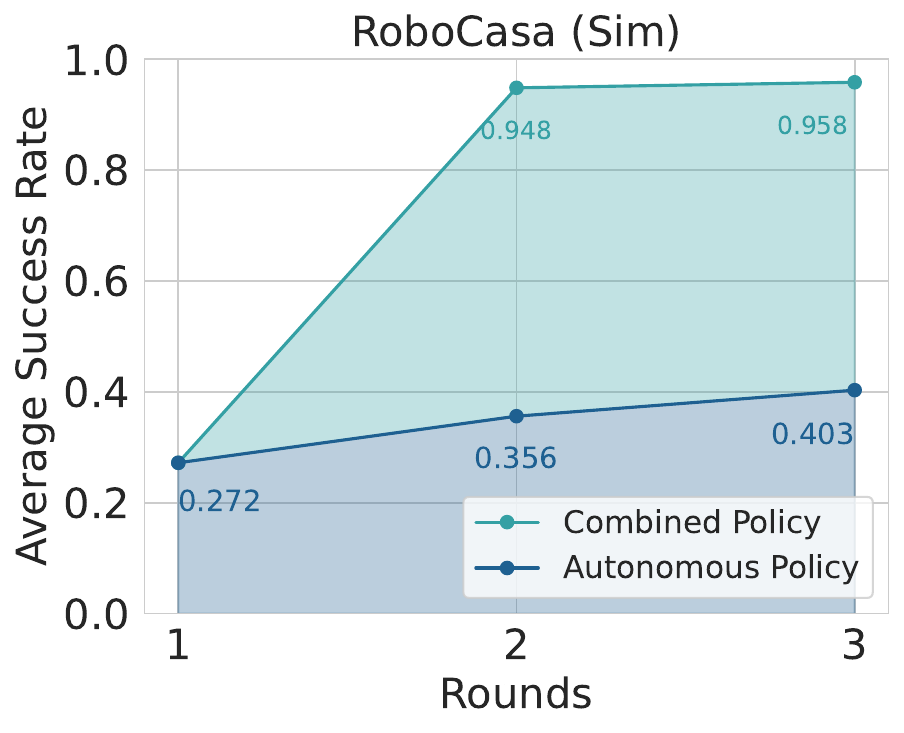}
    \end{subfigure}
    \hfill
    \begin{subfigure}[t]{0.305\textwidth}
        \centering
        \includegraphics[width=\textwidth]{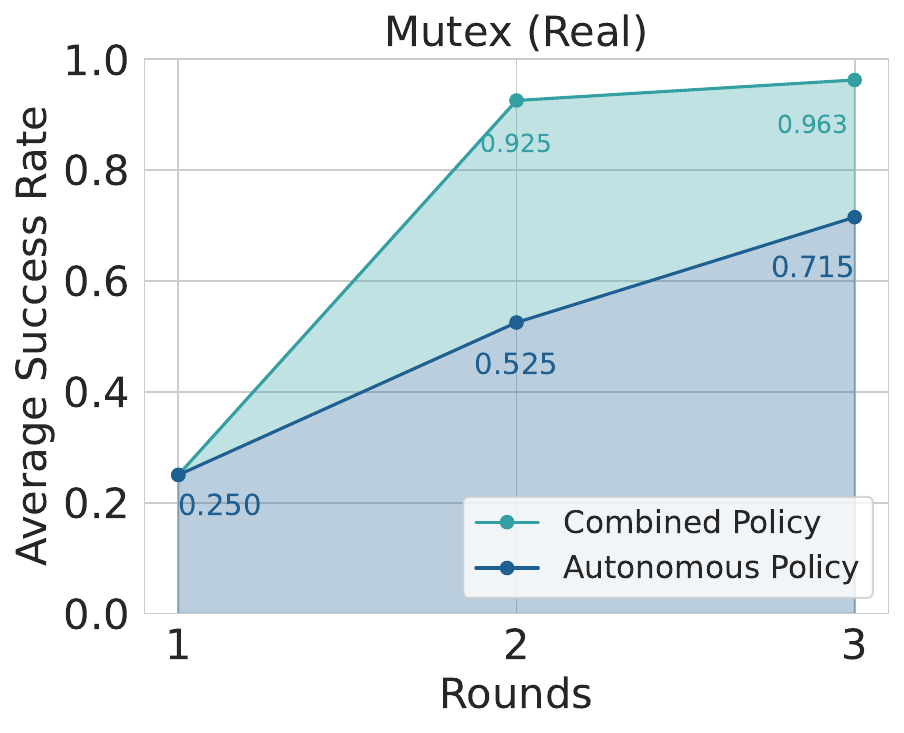}
    \end{subfigure}
    \hfill
    \begin{subfigure}[t]{0.18\textwidth}
        \centering
        \includegraphics[width=\textwidth]{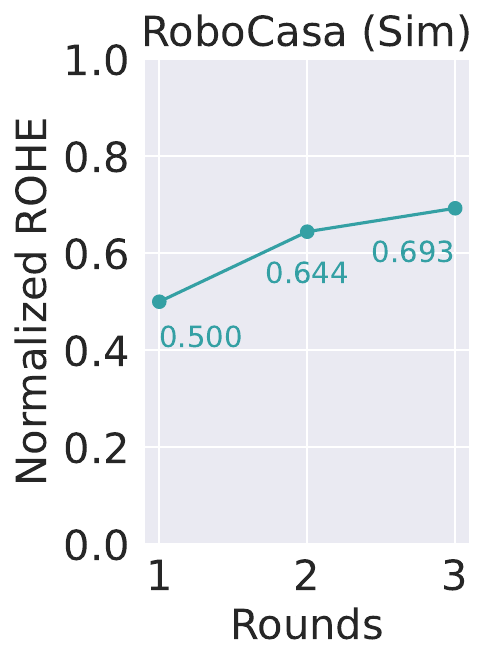}
    \end{subfigure}
    \hfill
    \begin{subfigure}[t]{0.18\textwidth}
        \centering
        \includegraphics[width=\textwidth]{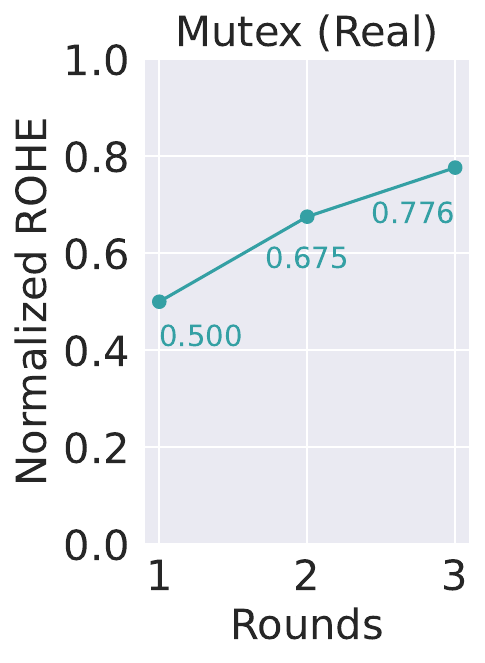}
    \end{subfigure}
    
    \caption{\textbf{System Performance Results.} \methodname{} shows consistent improvement throughout deployment for Combined and Autonomous Policy Performance and for Return of Human Effort (ROHE). The runtime monitoring ensures high combined policy performance and ROHE over time.}
    \vspace{-5mm}
    \label{fig:results}
\end{figure}

\subsection{Evaluation Results}

\noindent \textbf{System Performance Over Time.} Figure \ref{fig:results} presents the performance of \methodname{} over several deployment rounds. The autonomous policy improves consistently, showing a $13$\% increase in simulation and $45$\% in real-world settings. \methodname{} maintains high combined policy performance, exceeding $95$\% in both environments, demonstrating the effectiveness of runtime monitoring and timely human intervention. Additionally, \methodname{} shows improved ROHE performance over time. The continuous fine-tuning of anomaly predictors and adaptive thresholds leads to fewer predicted anomalies and reduced human intervention.

\begin{wrapfigure}{r}{0.5\textwidth}  
    \vspace{-4mm}
    \centering
    \includegraphics[width=0.5\textwidth]{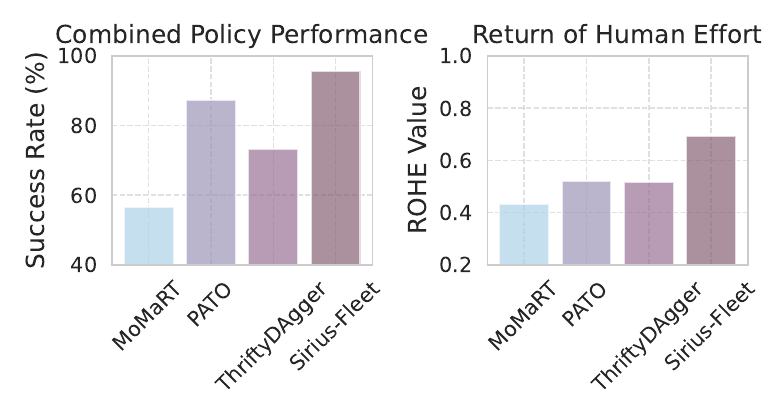}  
    \caption{\textbf{Baseline Comparison.} \methodname{} surpassed the baselines in both combined policy performance and ROHE.}
    \label{fig:baseline}
    \vspace{-4mm}
\end{wrapfigure}

\noindent \textbf{Baseline Comparison.} We compare \methodname{}'s multi-task runtime monitoring against three baseline methods: \textbf{MoMart} \cite{momart_wong22a}, \textbf{PATO} \cite{dass2022pato}, and \textbf{ThriftyDAgger} \cite{hoque2021thriftydagger}. MoMart uses VAE reconstruction loss for OOD detection, PATO combines ensemble policy variance with VAE reconstruction for predicting future image goals, and ThriftyDAgger merges OOD detection and failure detection using risky Q-function values.

Since these baselines are designed for single-task environments, we evaluated them on five tasks from different categories in RoboCasa, training separate models for each task. All baselines used fixed detection thresholds, while ThriftyDAgger adjusts thresholds based on a target intervention ratio. For fairness, we applied the same intervention ratio across all comparisons. As shown in Figure \ref{fig:baseline}, \methodname{} consistently outperforms the baselines in policy performance and ROHE. The fixed thresholds used by the baselines were ineffective in handling varying task distributions, resulting in excessive interventions or decreased performance across tasks.

\noindent \textbf{Ablations.} 
We conducted ablation studies to explore the following key questions: 1) How important is the design of the multi-task visual world model, as opposed to the single-task design? 2) What are the individual impacts of the Failure Prediction and OOD Prediction components? 3) How critical is the design of the multi-task policy? The results of these studies are detailed in Appendix \ref{appendix: ablation}.

\section{Conclusion}

We introduce \methodname{}, a framework for multi-task interactive robot fleet learning that combines a multi-task policy with runtime monitoring. Driven by a visual world model and task-adaptive anomaly predictors, \methodname{} improves policy performance and reduces human intervention through timely anomaly prediction.

\textbf{Limitations.} \methodname{} is best suited for quasi-static manipulation tasks, where anomalies can be easily corrected with teleoperation. However, applying \methodname{} to dynamic tasks could be challenging. Also, our experiments involve a small group of five human operators; future work could expand this by conducting large-scale human studies to understand the effects of diverse human interventions on runtime monitoring and policy training. Lastly, our evaluations used a single embodiment (Franka Emika Panda arm). Future research can extend \methodname{} to cross-embodiment fleet learning, making it more generalizable across different robot platforms \cite{open_x_embodiment_rt_x_2023}.

\acknowledgments{
We thank Zhenyu Jiang, Rutav Shah, Yifeng Zhu, and Shuijing Liu for providing helpful feedback on this manuscript. We also thank Xixi Hu, Zhenyu Jiang, Bo Liu, Yue Zhao, and Lizhang Chen for their fruitful discussions. We thank Soroush Nasiriany, Abhishek Joshi, and the RoboCasa team for their support in the simulation environments and rendering. This work was
partially supported by the National Science Foundation
(FRR2145283, EFRI-2318065), the Office of Naval Research
(N00014-22-1-2204, N00014-24-1-2550), and the DARPA
TIAMAT program (HR0011-24-9-0428).
}

\bibliography{example}  

\begin{thebibliography}{62}
\providecommand{\natexlab}[1]{#1}
\providecommand{\url}[1]{\texttt{#1}}
\expandafter\ifx\csname urlstyle\endcsname\relax
  \providecommand{\doi}[1]{doi: #1}\else
  \providecommand{\doi}{doi: \begingroup \urlstyle{rm}\Url}\fi

\bibitem[Brohan et~al.(2022)Brohan, Brown, Carbajal, Chebotar, Dabis, Finn, Gopalakrishnan, Hausman, Herzog, Hsu, Ibarz, Ichter, Irpan, Jackson, Jesmonth, Joshi, Julian, Kalashnikov, Kuang, Leal, Lee, Levine, Lu, Malla, Manjunath, Mordatch, Nachum, Parada, Peralta, Perez, Pertsch, Quiambao, Rao, Ryoo, Salazar, Sanketi, Sayed, Singh, Sontakke, Stone, Tan, Tran, Vanhoucke, Vega, Vuong, Xia, Xiao, Xu, Xu, Yu, and Zitkovich]{brohan2022rt1}
A.~Brohan, N.~Brown, J.~Carbajal, Y.~Chebotar, J.~Dabis, C.~Finn, K.~Gopalakrishnan, K.~Hausman, A.~Herzog, J.~Hsu, J.~Ibarz, B.~Ichter, A.~Irpan, T.~Jackson, S.~Jesmonth, N.~J. Joshi, R.~Julian, D.~Kalashnikov, Y.~Kuang, I.~Leal, K.-H. Lee, S.~Levine, Y.~Lu, U.~Malla, D.~Manjunath, I.~Mordatch, O.~Nachum, C.~Parada, J.~Peralta, E.~Perez, K.~Pertsch, J.~Quiambao, K.~Rao, M.~Ryoo, G.~Salazar, P.~Sanketi, K.~Sayed, J.~Singh, S.~Sontakke, A.~Stone, C.~Tan, H.~Tran, V.~Vanhoucke, S.~Vega, Q.~Vuong, F.~Xia, T.~Xiao, P.~Xu, S.~Xu, T.~Yu, and B.~Zitkovich.
\newblock Rt-1: Robotics transformer for real-world control at scale, 2022.

\bibitem[Reed et~al.(2022)Reed, Zolna, Parisotto, Colmenarejo, Novikov, Barth-Maron, Gimenez, Sulsky, Kay, Springenberg, Eccles, Bruce, Razavi, Edwards, Heess, Chen, Hadsell, Vinyals, Bordbar, and de~Freitas]{reed2022generalist}
S.~Reed, K.~Zolna, E.~Parisotto, S.~G. Colmenarejo, A.~Novikov, G.~Barth-Maron, M.~Gimenez, Y.~Sulsky, J.~Kay, J.~T. Springenberg, T.~Eccles, J.~Bruce, A.~Razavi, A.~Edwards, N.~Heess, Y.~Chen, R.~Hadsell, O.~Vinyals, M.~Bordbar, and N.~de~Freitas.
\newblock A generalist agent, 2022.

\bibitem[Shah et~al.(2023)Shah, Sridhar, Dashora, Stachowicz, Black, Hirose, and Levine]{shah2023vint}
D.~Shah, A.~Sridhar, N.~Dashora, K.~Stachowicz, K.~Black, N.~Hirose, and S.~Levine.
\newblock Vint: A foundation model for visual navigation, 2023.

\bibitem[Hoque et~al.(2023)Hoque, Chen, Sharma, Dharmarajan, Thananjeyan, Abbeel, and Goldberg]{hoque2023fleet}
R.~Hoque, L.~Y. Chen, S.~Sharma, K.~Dharmarajan, B.~Thananjeyan, P.~Abbeel, and K.~Goldberg.
\newblock Fleet-dagger: Interactive robot fleet learning with scalable human supervision.
\newblock In \emph{Conference on Robot Learning}, pages 368--380. PMLR, 2023.

\bibitem[Wang et~al.(2024)Wang, Zhang, Zhou, Simchowitz, and Tedrake]{wang2024robot}
L.~Wang, K.~Zhang, A.~Zhou, M.~Simchowitz, and R.~Tedrake.
\newblock Robot fleet learning via policy merging, 2024.

\bibitem[Müller et~al.(2024)Müller, Brahmbhatt, Deka, Leboutet, Hafner, and Koltun]{müller2024openbotfleet}
M.~Müller, S.~Brahmbhatt, A.~Deka, Q.~Leboutet, D.~Hafner, and V.~Koltun.
\newblock Openbot-fleet: A system for collective learning with real robots, 2024.

\bibitem[Mandlekar et~al.(2020)Mandlekar, Xu, Martín-Martín, Zhu, Fei-Fei, and Savarese]{mandlekar2020humanintheloop}
A.~Mandlekar, D.~Xu, R.~Martín-Martín, Y.~Zhu, L.~Fei-Fei, and S.~Savarese.
\newblock Human-in-the-loop imitation learning using remote teleoperation.
\newblock In \emph{arXiv preprint arXiv:2012.06733}, 2020.

\bibitem[Chisari et~al.(2021)Chisari, Welschehold, Boedecker, Burgard, and Valada]{2022correct_me}
E.~Chisari, T.~Welschehold, J.~Boedecker, W.~Burgard, and A.~Valada.
\newblock Correct me if i am wrong: Interactive learning for robotic manipulation.
\newblock In \emph{RAL}, volume~7, pages 3695--3702, 2021.

\bibitem[Spencer et~al.(2020)Spencer, Choudhury, Barnes, Schmittle, Chiang, Ramadge, and Srinivasa]{spencer2020learning}
J.~Spencer, S.~Choudhury, M.~Barnes, M.~Schmittle, M.~Chiang, P.~Ramadge, and S.~Srinivasa.
\newblock Learning from interventions: Human-robot interaction as both explicit and implicit feedback.
\newblock In \emph{16th Robotics: Science and Systems, RSS 2020}. MIT Press Journals, 2020.

\bibitem[Kelly et~al.(2019)Kelly, Sidrane, Driggs-Campbell, and Kochenderfer]{hgdagger2019}
M.~Kelly, C.~Sidrane, K.~Driggs-Campbell, and M.~J. Kochenderfer.
\newblock Hg-dagger: Interactive imitation learning with human experts.
\newblock \emph{2019 International Conference on Robotics and Automation (ICRA)}, May 2019.
\newblock \doi{10.1109/icra.2019.8793698}.
\newblock URL \url{http://dx.doi.org/10.1109/ICRA.2019.8793698}.

\bibitem[Hoque et~al.(2021{\natexlab{a}})Hoque, Balakrishna, Putterman, Luo, Brown, Seita, Thananjeyan, Novoseller, and Goldberg]{hoque2021lazydagger}
R.~Hoque, A.~Balakrishna, C.~Putterman, M.~Luo, D.~S. Brown, D.~Seita, B.~Thananjeyan, E.~Novoseller, and K.~Goldberg.
\newblock Lazydagger: Reducing context switching in interactive imitation learning.
\newblock In \emph{2021 IEEE 17th International Conference on Automation Science and Engineering (CASE)}, pages 502--509. IEEE, 2021{\natexlab{a}}.

\bibitem[Hoque et~al.(2021{\natexlab{b}})Hoque, Balakrishna, Novoseller, Wilcox, Brown, and Goldberg]{hoque2021thriftydagger}
R.~Hoque, A.~Balakrishna, E.~Novoseller, A.~Wilcox, D.~S. Brown, and K.~Goldberg.
\newblock Thriftydagger: Budget-aware novelty and risk gating for interactive imitation learning.
\newblock \emph{arXiv preprint arXiv:2109.08273}, 2021{\natexlab{b}}.

\bibitem[Liu et~al.(2023)Liu, Nasiriany, Zhang, Bao, and Zhu]{liu2022robot}
H.~Liu, S.~Nasiriany, L.~Zhang, Z.~Bao, and Y.~Zhu.
\newblock Robot learning on the job: Human-in-the-loop autonomy and learning during deployment.
\newblock In \emph{Robotics: Science and Systems (RSS)}, 2023.

\bibitem[Hoque et~al.(2022)Hoque, Chen, Sharma, Dharmarajan, Thananjeyan, Abbeel, and Goldberg]{hoque2022fleetdagger}
R.~Hoque, L.~Y. Chen, S.~Sharma, K.~Dharmarajan, B.~Thananjeyan, P.~Abbeel, and K.~Goldberg.
\newblock Fleet-dagger: Interactive robot fleet learning with scalable human supervision, 2022.

\bibitem[Li et~al.(2022)Li, Peng, and Zhou]{li2022efficient}
Q.~Li, Z.~Peng, and B.~Zhou.
\newblock Efficient learning of safe driving policy via human-ai copilot optimization, 2022.

\bibitem[Yel and Bezzo(2019)]{yel_runtime}
E.~Yel and N.~Bezzo.
\newblock Fast run-time monitoring, replanning, and recovery for safe autonomous system operations.
\newblock In \emph{2019 IEEE/RSJ International Conference on Intelligent Robots and Systems (IROS)}, pages 1661--1667, 2019.
\newblock \doi{10.1109/IROS40897.2019.8968498}.

\bibitem[Liu et~al.(2024)Liu, Dass, Martín-Martín, and Zhu]{liu2023modelbased}
H.~Liu, S.~Dass, R.~Martín-Martín, and Y.~Zhu.
\newblock Model-based runtime monitoring with interactive imitation learning.
\newblock In \emph{IEEE International Conference on Robotics and Automation (ICRA)}, 2024.

\bibitem[Salehi et~al.(2021)Salehi, Mirzaei, Hendrycks, Li, Rohban, and Sabokrou]{ood_survey}
M.~Salehi, H.~Mirzaei, D.~Hendrycks, Y.~Li, M.~H. Rohban, and M.~Sabokrou.
\newblock A unified survey on anomaly, novelty, open-set, and out-of-distribution detection: Solutions and future challenges.
\newblock \emph{CoRR}, abs/2110.14051, 2021.
\newblock URL \url{https://arxiv.org/abs/2110.14051}.

\bibitem[Richter and Roy(2017)]{Richter_OOD}
C.~Richter and N.~Roy.
\newblock Safe visual navigation via deep learning and novelty detection.
\newblock \emph{Robotics: Science and Systems}, 07 2017.

\bibitem[Wong et~al.(2022)Wong, Tung, Kurenkov, Mandlekar, Fei-Fei, Savarese, and Mart{\'i}n-Mart{\'i}n]{momart_wong22a}
J.~Wong, A.~Tung, A.~Kurenkov, A.~Mandlekar, L.~Fei-Fei, S.~Savarese, and R.~Mart{\'i}n-Mart{\'i}n.
\newblock Error-aware imitation learning from teleoperation data for mobile manipulation.
\newblock In \emph{Proceedings of the 5th Conference on Robot Learning}, volume 164 of \emph{Proceedings of Machine Learning Research}, pages 1367--1378, 2022.

\bibitem[Dass et~al.(2022)Dass, Pertsch, Zhang, Lee, Lim, and Nikolaidis]{dass2022pato}
S.~Dass, K.~Pertsch, H.~Zhang, Y.~Lee, J.~J. Lim, and S.~Nikolaidis.
\newblock Pato: Policy assisted teleoperation for scalable robot data collection, 2022.

\bibitem[Brooks et~al.(2024)Brooks, Peebles, Holmes, DePue, Guo, Jing, Schnurr, Taylor, Luhman, Luhman, Ng, Wang, and Ramesh]{videoworldsimulators2024}
T.~Brooks, B.~Peebles, C.~Holmes, W.~DePue, Y.~Guo, L.~Jing, D.~Schnurr, J.~Taylor, T.~Luhman, E.~Luhman, C.~Ng, R.~Wang, and A.~Ramesh.
\newblock Video generation models as world simulators.
\newblock 2024.

\bibitem[Bruce et~al.(2024)Bruce, Dennis, Edwards, Parker-Holder, Shi, Hughes, Lai, Mavalankar, Steigerwald, Apps, Aytar, Bechtle, Behbahani, Chan, Heess, Gonzalez, Osindero, Ozair, Reed, Zhang, Zolna, Clune, de~Freitas, Singh, and Rocktäschel]{bruce2024genie}
J.~Bruce, M.~Dennis, A.~Edwards, J.~Parker-Holder, Y.~Shi, E.~Hughes, M.~Lai, A.~Mavalankar, R.~Steigerwald, C.~Apps, Y.~Aytar, S.~Bechtle, F.~Behbahani, S.~Chan, N.~Heess, L.~Gonzalez, S.~Osindero, S.~Ozair, S.~Reed, J.~Zhang, K.~Zolna, J.~Clune, N.~de~Freitas, S.~Singh, and T.~Rocktäschel.
\newblock Genie: Generative interactive environments, 2024.

\bibitem[Xiang et~al.(2024)Xiang, Liu, Gu, Gao, Ning, Zha, Feng, Tao, Hao, Shi, Liu, Xing, and Hu]{xiang2024pandora}
J.~Xiang, G.~Liu, Y.~Gu, Q.~Gao, Y.~Ning, Y.~Zha, Z.~Feng, T.~Tao, S.~Hao, Y.~Shi, Z.~Liu, E.~P. Xing, and Z.~Hu.
\newblock Pandora: Towards general world model with natural language actions and video states.
\newblock 2024.

\bibitem[Hu et~al.(2023)Hu, Russell, Yeo, Murez, Fedoseev, Kendall, Shotton, and Corrado]{hu2023gaia1}
A.~Hu, L.~Russell, H.~Yeo, Z.~Murez, G.~Fedoseev, A.~Kendall, J.~Shotton, and G.~Corrado.
\newblock Gaia-1: A generative world model for autonomous driving, 2023.

\bibitem[Jiang et~al.(2023)Jiang, Gupta, Zhang, Wang, Dou, Chen, Fei-Fei, Anandkumar, Zhu, and Fan]{jiang2023vima}
Y.~Jiang, A.~Gupta, Z.~Zhang, G.~Wang, Y.~Dou, Y.~Chen, L.~Fei-Fei, A.~Anandkumar, Y.~Zhu, and L.~Fan.
\newblock Vima: General robot manipulation with multimodal prompts.
\newblock In \emph{Fortieth International Conference on Machine Learning}, 2023.

\bibitem[Shah et~al.(2023)Shah, Mart{\'\i}n-Mart{\'\i}n, and Zhu]{shah2023mutex}
R.~Shah, R.~Mart{\'\i}n-Mart{\'\i}n, and Y.~Zhu.
\newblock Mutex: Learning unified policies from multimodal task specifications.
\newblock In \emph{7th Annual Conference on Robot Learning}, 2023.
\newblock URL \url{https://openreview.net/forum?id=PwqiqaaEzJ}.

\bibitem[Brohan et~al.(2023)Brohan, Brown, Carbajal, Chebotar, Chen, Choromanski, Ding, Driess, Dubey, Finn, Florence, Fu, Arenas, Gopalakrishnan, Han, Hausman, Herzog, Hsu, Ichter, Irpan, Joshi, Julian, Kalashnikov, Kuang, Leal, Lee, Lee, Levine, Lu, Michalewski, Mordatch, Pertsch, Rao, Reymann, Ryoo, Salazar, Sanketi, Sermanet, Singh, Singh, Soricut, Tran, Vanhoucke, Vuong, Wahid, Welker, Wohlhart, Wu, Xia, Xiao, Xu, Xu, Yu, and Zitkovich]{brohan2023rt2}
A.~Brohan, N.~Brown, J.~Carbajal, Y.~Chebotar, X.~Chen, K.~Choromanski, T.~Ding, D.~Driess, A.~Dubey, C.~Finn, P.~Florence, C.~Fu, M.~G. Arenas, K.~Gopalakrishnan, K.~Han, K.~Hausman, A.~Herzog, J.~Hsu, B.~Ichter, A.~Irpan, N.~Joshi, R.~Julian, D.~Kalashnikov, Y.~Kuang, I.~Leal, L.~Lee, T.-W.~E. Lee, S.~Levine, Y.~Lu, H.~Michalewski, I.~Mordatch, K.~Pertsch, K.~Rao, K.~Reymann, M.~Ryoo, G.~Salazar, P.~Sanketi, P.~Sermanet, J.~Singh, A.~Singh, R.~Soricut, H.~Tran, V.~Vanhoucke, Q.~Vuong, A.~Wahid, S.~Welker, P.~Wohlhart, J.~Wu, F.~Xia, T.~Xiao, P.~Xu, S.~Xu, T.~Yu, and B.~Zitkovich.
\newblock Rt-2: Vision-language-action models transfer web knowledge to robotic control, 2023.

\bibitem[Jang et~al.(2022)Jang, Irpan, Khansari, Kappler, Ebert, Lynch, Levine, and Finn]{jang2022bcz}
E.~Jang, A.~Irpan, M.~Khansari, D.~Kappler, F.~Ebert, C.~Lynch, S.~Levine, and C.~Finn.
\newblock Bc-z: Zero-shot task generalization with robotic imitation learning, 2022.

\bibitem[Bousmalis et~al.(2023)Bousmalis, Vezzani, Rao, Devin, Lee, Bauza, Davchev, Zhou, Gupta, Raju, Laurens, Fantacci, Dalibard, Zambelli, Martins, Pevceviciute, Blokzijl, Denil, Batchelor, Lampe, Parisotto, Żołna, Reed, Colmenarejo, Scholz, Abdolmaleki, Groth, Regli, Sushkov, Rothörl, Chen, Aytar, Barker, Ortiz, Riedmiller, Springenberg, Hadsell, Nori, and Heess]{bousmalis2023robocat}
K.~Bousmalis, G.~Vezzani, D.~Rao, C.~Devin, A.~X. Lee, M.~Bauza, T.~Davchev, Y.~Zhou, A.~Gupta, A.~Raju, A.~Laurens, C.~Fantacci, V.~Dalibard, M.~Zambelli, M.~Martins, R.~Pevceviciute, M.~Blokzijl, M.~Denil, N.~Batchelor, T.~Lampe, E.~Parisotto, K.~Żołna, S.~Reed, S.~G. Colmenarejo, J.~Scholz, A.~Abdolmaleki, O.~Groth, J.-B. Regli, O.~Sushkov, T.~Rothörl, J.~E. Chen, Y.~Aytar, D.~Barker, J.~Ortiz, M.~Riedmiller, J.~T. Springenberg, R.~Hadsell, F.~Nori, and N.~Heess.
\newblock Robocat: A self-improving generalist agent for robotic manipulation, 2023.

\bibitem[Team et~al.(2024)Team, Ghosh, Walke, Pertsch, Black, Mees, Dasari, Hejna, Kreiman, Xu, Luo, Tan, Chen, Sanketi, Vuong, Xiao, Sadigh, Finn, and Levine]{octomodelteam2024octo}
O.~M. Team, D.~Ghosh, H.~Walke, K.~Pertsch, K.~Black, O.~Mees, S.~Dasari, J.~Hejna, T.~Kreiman, C.~Xu, J.~Luo, Y.~L. Tan, L.~Y. Chen, P.~Sanketi, Q.~Vuong, T.~Xiao, D.~Sadigh, C.~Finn, and S.~Levine.
\newblock Octo: An open-source generalist robot policy, 2024.

\bibitem[Mandlekar et~al.(2021)Mandlekar, Xu, Wong, Nasiriany, Wang, Kulkarni, Fei-Fei, Savarese, Zhu, and Martín-Martín]{robomimic}
A.~Mandlekar, D.~Xu, J.~Wong, S.~Nasiriany, C.~Wang, R.~Kulkarni, L.~Fei-Fei, S.~Savarese, Y.~Zhu, and R.~Martín-Martín.
\newblock What matters in learning from offline human demonstrations for robot manipulation, 2021.

\bibitem[Ho et~al.(2020)Ho, Jain, and Abbeel]{ho2020denoising}
J.~Ho, A.~Jain, and P.~Abbeel.
\newblock Denoising diffusion probabilistic models, 2020.

\bibitem[Chi et~al.(2024)Chi, Xu, Feng, Cousineau, Du, Burchfiel, Tedrake, and Song]{chi2024diffusion}
C.~Chi, Z.~Xu, S.~Feng, E.~Cousineau, Y.~Du, B.~Burchfiel, R.~Tedrake, and S.~Song.
\newblock Diffusion policy: Visuomotor policy learning via action diffusion, 2024.

\bibitem[Ebert et~al.(2021)Ebert, Yang, Schmeckpeper, Bucher, Georgakis, Daniilidis, Finn, and Levine]{ebert2021bridge}
F.~Ebert, Y.~Yang, K.~Schmeckpeper, B.~Bucher, G.~Georgakis, K.~Daniilidis, C.~Finn, and S.~Levine.
\newblock Bridge data: Boosting generalization of robotic skills with cross-domain datasets, 2021.

\bibitem[{Open X-Embodiment Collaboration} et~al.(2024)]{open_x_embodiment_rt_x_2023}
{Open X-Embodiment Collaboration} et~al.
\newblock Open {X-E}mbodiment: Robotic learning datasets and {RT-X} models.
\newblock In \emph{Proceedings of the IEEE International Conference on Robotics and Automation (ICRA)}, Yokohama, Japan, 2024.

\bibitem[Khazatsky et~al.(2024)Khazatsky, Pertsch, Nair, Balakrishna, Dasari, Karamcheti, Nasiriany, Srirama, Chen, et~al.]{khazatsky2024droid}
A.~Khazatsky, K.~Pertsch, S.~Nair, A.~Balakrishna, S.~Dasari, S.~Karamcheti, S.~Nasiriany, M.~K. Srirama, L.~Y. Chen, et~al.
\newblock Droid: A large-scale in-the-wild robot manipulation dataset, 2024.

\bibitem[Mandlekar et~al.(2023)Mandlekar, Nasiriany, Wen, Akinola, Narang, Fan, Zhu, and Fox]{mandlekar2023mimicgen}
A.~Mandlekar, S.~Nasiriany, B.~Wen, I.~Akinola, Y.~Narang, L.~Fan, Y.~Zhu, and D.~Fox.
\newblock Mimicgen: A data generation system for scalable robot learning using human demonstrations.
\newblock In \emph{7th Annual Conference on Robot Learning}, 2023.

\bibitem[Nasiriany et~al.(2024)Nasiriany, Maddukuri, Zhang, Parikh, Lo, Joshi, Mandlekar, and Zhu]{robocasa2024}
S.~Nasiriany, A.~Maddukuri, L.~Zhang, A.~Parikh, A.~Lo, A.~Joshi, A.~Mandlekar, and Y.~Zhu.
\newblock Robocasa: Large-scale simulation of everyday tasks for generalist robots.
\newblock In \emph{Robotics: Science and Systems (RSS)}, 2024.

\bibitem[Kalashnikov et~al.(2018)Kalashnikov, Irpan, Pastor, Ibarz, Herzog, Jang, Quillen, Holly, Kalakrishnan, Vanhoucke, and Levine]{kalashnikov2018qt}
D.~Kalashnikov, A.~Irpan, P.~Pastor, J.~Ibarz, A.~Herzog, E.~Jang, D.~Quillen, E.~Holly, M.~Kalakrishnan, V.~Vanhoucke, and S.~Levine.
\newblock {QT-Opt}: Scalable deep reinforcement learning for vision-based robotic manipulation.
\newblock In \emph{CoRL}, 2018.

\bibitem[Team et~al.(2024)Team, Aldaco, Armstrong, Baruch, Bingham, Chan, Draper, Dwibedi, Finn, Florence, Goodrich, Gramlich, Hage, Herzog, Hoech, Nguyen, Storz, Tabanpour, Takayama, Tompson, Wahid, Wahrburg, Xu, Yaroshenko, Zakka, and Zhao]{aloha2team2024aloha}
A.~. Team, J.~Aldaco, T.~Armstrong, R.~Baruch, J.~Bingham, S.~Chan, K.~Draper, D.~Dwibedi, C.~Finn, P.~Florence, S.~Goodrich, W.~Gramlich, T.~Hage, A.~Herzog, J.~Hoech, T.~Nguyen, I.~Storz, B.~Tabanpour, L.~Takayama, J.~Tompson, A.~Wahid, T.~Wahrburg, S.~Xu, S.~Yaroshenko, K.~Zakka, and T.~Z. Zhao.
\newblock Aloha 2: An enhanced low-cost hardware for bimanual teleoperation, 2024.

\bibitem[Ahn et~al.(2024)Ahn, Dwibedi, Finn, Arenas, Gopalakrishnan, Hausman, Ichter, Irpan, Joshi, Julian, Kirmani, Leal, Lee, Levine, Lu, Leal, Maddineni, Rao, Sadigh, Sanketi, Sermanet, Vuong, Welker, Xia, Xiao, Xu, Xu, and Xu]{ahn2024autort}
M.~Ahn, D.~Dwibedi, C.~Finn, M.~G. Arenas, K.~Gopalakrishnan, K.~Hausman, B.~Ichter, A.~Irpan, N.~Joshi, R.~Julian, S.~Kirmani, I.~Leal, E.~Lee, S.~Levine, Y.~Lu, I.~Leal, S.~Maddineni, K.~Rao, D.~Sadigh, P.~Sanketi, P.~Sermanet, Q.~Vuong, S.~Welker, F.~Xia, T.~Xiao, P.~Xu, S.~Xu, and Z.~Xu.
\newblock Autort: Embodied foundation models for large scale orchestration of robotic agents, 2024.

\bibitem[Swamy et~al.(2020)Swamy, Reddy, Levine, and Dragan]{swamy2020scaled}
G.~Swamy, S.~Reddy, S.~Levine, and A.~D. Dragan.
\newblock Scaled autonomy: Enabling human operators to control robot fleets, 2020.

\bibitem[Akcin et~al.(2023{\natexlab{a}})Akcin, Li, Agarwal, and Chinchali]{akcin23a_decentral}
O.~Akcin, P.-h. Li, S.~Agarwal, and S.~P. Chinchali.
\newblock Decentralized data collection for robotic fleet learning: A game-theoretic approach.
\newblock In K.~Liu, D.~Kulic, and J.~Ichnowski, editors, \emph{Proceedings of The 6th Conference on Robot Learning}, volume 205 of \emph{Proceedings of Machine Learning Research}, pages 978--988. PMLR, 14--18 Dec 2023{\natexlab{a}}.
\newblock URL \url{https://proceedings.mlr.press/v205/akcin23a.html}.

\bibitem[Akcin et~al.(2023{\natexlab{b}})Akcin, Unuvar, Ure, and Chinchali]{akcin23a_fleet_active}
O.~Akcin, O.~Unuvar, O.~Ure, and S.~P. Chinchali.
\newblock Fleet active learning: A submodular maximization approach.
\newblock In J.~Tan, M.~Toussaint, and K.~Darvish, editors, \emph{Proceedings of The 7th Conference on Robot Learning}, volume 229 of \emph{Proceedings of Machine Learning Research}, pages 1378--1399. PMLR, 06--09 Nov 2023{\natexlab{b}}.
\newblock URL \url{https://proceedings.mlr.press/v229/akcin23a.html}.

\bibitem[Hsu et~al.(2023)Hsu, Hu, and Fisac]{Hsu2023TheSF}
K.-C. Hsu, H.~Hu, and J.~F. Fisac.
\newblock The safety filter: A unified view of safety-critical control in autonomous systems.
\newblock 2023.
\newblock URL \url{https://api.semanticscholar.org/CorpusID:261697421}.

\bibitem[Sinha et~al.(2023)Sinha, Schmerling, and Pavone]{SinhaSchmerlingEtAl2023}
R.~Sinha, E.~Schmerling, and M.~Pavone.
\newblock Closing the loop on runtime monitors with fallback-safe mpc.
\newblock In \emph{{Proc. IEEE Conf. on Decision and Control}}, 2023.

\bibitem[Nakamura et~al.(2024)Nakamura, Tian, and Bajcsy]{Nakamura2024AGC}
K.~Nakamura, R.~Tian, and A.~Bajcsy.
\newblock Not all errors are made equal: A regret metric for detecting system-level trajectory prediction failures.
\newblock In \emph{8th Conference on Robot Learning}, 2024.

\bibitem[Sinha et~al.(2022)Sinha, Sharma, Banerjee, Lew, Luo, Richards, Sun, Schmerling, and Pavone]{sinha2022systemlevel}
R.~Sinha, A.~Sharma, S.~Banerjee, T.~Lew, R.~Luo, S.~M. Richards, Y.~Sun, E.~Schmerling, and M.~Pavone.
\newblock A system-level view on out-of-distribution data in robotics, 2022.

\bibitem[Menda et~al.(2019)Menda, Driggs-Campbell, and Kochenderfer]{menda2019ensembledagger}
K.~Menda, K.~Driggs-Campbell, and M.~J. Kochenderfer.
\newblock {EnsembleDagger}: A bayesian approach to safe imitation learning.
\newblock In \emph{2019 IEEE/RSJ International Conference on Intelligent Robots and Systems (IROS)}, pages 5041--5048. IEEE, 2019.

\bibitem[Diryag et~al.(2014)Diryag, Miti{\'c}, and Miljkovi{\'c}]{diryag2014neural}
A.~Diryag, M.~Miti{\'c}, and Z.~Miljkovi{\'c}.
\newblock Neural networks for prediction of robot failures.
\newblock \emph{Proceedings of the Institution of Mechanical Engineers, Part C: Journal of Mechanical Engineering Science}, 228\penalty0 (8):\penalty0 1444--1458, 2014.

\bibitem[Gokmen et~al.(2023)Gokmen, Ho, and Khansari]{Gokmen2023AskingFH}
C.~Gokmen, D.~Ho, and M.~Khansari.
\newblock Asking for help: Failure prediction in behavioral cloning through value approximation.
\newblock \emph{ArXiv}, abs/2302.04334, 2023.

\bibitem[Xie et~al.(2022)Xie, Tajwar, Sharma, and Finn]{xie2022ask}
A.~Xie, F.~Tajwar, A.~Sharma, and C.~Finn.
\newblock When to ask for help: Proactive interventions in autonomous reinforcement learning, 2022.

\bibitem[Ren et~al.(2023)Ren, Dixit, Bodrova, Singh, Tu, Brown, Xu, Takayama, Xia, Varley, Xu, Sadigh, Zeng, and Majumdar]{ren2023robots}
A.~Z. Ren, A.~Dixit, A.~Bodrova, S.~Singh, S.~Tu, N.~Brown, P.~Xu, L.~Takayama, F.~Xia, J.~Varley, Z.~Xu, D.~Sadigh, A.~Zeng, and A.~Majumdar.
\newblock Robots that ask for help: Uncertainty alignment for large language model planners, 2023.

\bibitem[Liu et~al.(2023)Liu, Bahety, and Song]{liu2023reflect}
Z.~Liu, A.~Bahety, and S.~Song.
\newblock Reflect: Summarizing robot experiences for failure explanation and correction, 2023.

\bibitem[Guan et~al.(2024)Guan, Zhou, Liu, Zha, Amor, and Kambhampati]{guan2024task}
L.~Guan, Y.~Zhou, D.~Liu, Y.~Zha, H.~B. Amor, and S.~Kambhampati.
\newblock "task success" is not enough: Investigating the use of video-language models as behavior critics for catching undesirable agent behaviors, 2024.

\bibitem[Hu et~al.(2022)Hu, Corrado, Griffiths, Murez, Gurau, Yeo, Kendall, Cipolla, and Shotton]{hu2022modelbased}
A.~Hu, G.~Corrado, N.~Griffiths, Z.~Murez, C.~Gurau, H.~Yeo, A.~Kendall, R.~Cipolla, and J.~Shotton.
\newblock Model-based imitation learning for urban driving, 2022.

\bibitem[Ronneberger et~al.(2015)Ronneberger, Fischer, and Brox]{ronneberger2015unet}
O.~Ronneberger, P.~Fischer, and T.~Brox.
\newblock U-net: Convolutional networks for biomedical image segmentation, 2015.

\bibitem[Sohn et~al.(2015)Sohn, Lee, and Yan]{NIPS2015_8d55a249}
K.~Sohn, H.~Lee, and X.~Yan.
\newblock Learning structured output representation using deep conditional generative models.
\newblock In C.~Cortes, N.~Lawrence, D.~Lee, M.~Sugiyama, and R.~Garnett, editors, \emph{Advances in Neural Information Processing Systems}, volume~28. Curran Associates, Inc., 2015.
\newblock URL \url{https://proceedings.neurips.cc/paper_files/paper/2015/file/8d55a249e6baa5c06772297520da2051-Paper.pdf}.

\bibitem[Wu et~al.(2023)Wu, Escontrela, Hafner, Abbeel, and Goldberg]{wu2023daydreamer}
P.~Wu, A.~Escontrela, D.~Hafner, P.~Abbeel, and K.~Goldberg.
\newblock Daydreamer: World models for physical robot learning.
\newblock In \emph{Conference on Robot Learning}, pages 2226--2240. PMLR, 2023.

\bibitem[Vaswani et~al.(2023)Vaswani, Shazeer, Parmar, Uszkoreit, Jones, Gomez, Kaiser, and Polosukhin]{vaswani2023attention}
A.~Vaswani, N.~Shazeer, N.~Parmar, J.~Uszkoreit, L.~Jones, A.~N. Gomez, L.~Kaiser, and I.~Polosukhin.
\newblock Attention is all you need, 2023.

\bibitem[Mandlekar et~al.(2018)Mandlekar, Zhu, Garg, Booher, Spero, Tung, Gao, Emmons, Gupta, Orbay, Savarese, and Fei-Fei]{mandlekar2018roboturk}
A.~Mandlekar, Y.~Zhu, A.~Garg, J.~Booher, M.~Spero, A.~Tung, J.~Gao, J.~Emmons, A.~Gupta, E.~Orbay, S.~Savarese, and L.~Fei-Fei.
\newblock Roboturk: A crowdsourcing platform for robotic skill learning through imitation.
\newblock In \emph{CoRL}, pages 879--893, 2018.

\end{thebibliography}

\clearpage
\section{Appendix}

\subsection{Table of Contents}

\begin{itemize}
    \item Ablations (Section \ref{appendix: ablation}) 
    \item Implementation Details (Section \ref{appendix: impl-details})
    \item Qualitative Analysis and Discussion (Section \ref{appendix: qual})
    \item Additional Details on Tasks (Section \ref{appendix: tasks})
\end{itemize}

\subsection{Ablations}
\label{appendix: ablation}

\textit{Multi-Task World Model vs. Single-Task World Model.}  
We empirically show how the multi-task world model is more generalizable than the single-task version in terms of future latent state prediction accuracy. Accurate future latent state prediction is essential for effective error prediction, as the downstream error predictors directly use the predicted future states. We compared the performance of \methodname{}'s multi-task world model against six single-task world models in terms of mean squared error (MSE) for future latent state prediction. As shown in Table \ref{tab:single-multi-task}, the multi-task world model consistently outperforms the single-task models across all tasks. This indicates that the multi-task world model is crucial for providing more accurate state predictions. 

\textit{Ablation of Failure and OOD Prediction.} We examine the impact of OOD and failure prediction components on human intervention overlap accuracy: We have a human operator fully supervise the robot execution of a policy and can intervene whenever an unsafe state is observed. We then apply the learned error predictors to the collected trajectories and compare failure states identified by the error predictors with those intervened by the human operator. This shows how the error predicted aligns with humans' risk assessment.

As shown in Table \ref{tab: ablation-ood-failure}, the combined use of OOD prediction and failure prediction by \methodname{} outperforms using either component alone across all tasks. This indicates that both accurate OOD and failure prediction are essential to the overall policy performance, as they work together to identify potential errors during deployment. Videos of trajectories predicted as OOD or failures are available on the project’s \href{https://ut-austin-rpl.github.io/sirius-fleet/}{website}.

\textit{Multi-task Policy vs. Single-task Policy.} We show that training on multi-task settings improves the overall performance of robot policy than training on individual single tasks. We show the policy success rate (\%) for 7 different tasks in Figure \ref{fig:multi-single-policy}, where we compare the multi-task policy trained on all tasks with single-task policies trained on individual tasks. For most tasks, \methodname{}'s multi-task policy performs better than the single-task policy trained on that specific task. This aligns with the observation in prior works \cite{octomodelteam2024octo, open_x_embodiment_rt_x_2023, reed2022generalist} that multi-task training helps the policy generalize better.
\begin{table}[h!]
\centering
\begin{tabular}{l|c|c}
\toprule
\textbf{Task} & \textbf{Single Task} & \textbf{Multi Task} \\
\midrule
\texttt{CloseDoorSingleHinge} & \num{3.3e-4} & \textbf{\num{1.5e-4}} \\
\texttt{PnPCabToCounter}     & \num{3.1e-4} & \textbf{\num{1.7e-4}} \\
\texttt{TurnOnMicrowave}      & \num{1.7e-4} & \textbf{\num{1.3e-4}} \\
\texttt{TurnOnSinkFaucet}     & \num{3.2e-4} & \textbf{\num{1.2e-4}} \\
\texttt{CoffeeSetupMug}       & \num{4.4e-4} & \textbf{\num{1.8e-4}} \\
\bottomrule
\end{tabular}
\vspace{2mm}
\caption{\textbf{Multi-task World Model vs. Single-task World Model Comparison.} We show the MSE loss ($\downarrow$) for future latent states prediction, where \methodname{} 's multi-task world model is compared with 6 different world models trained on 6 single tasks. The multi-task world model consistently gives more accurate future latent prediction than the world models trained with single-task data. }
\label{tab:single-multi-task}
\end{table}

\begin{table}[h!]
\centering
\begin{tabular}{lccc}
\toprule
Task & \methodname{} & \methodname{} & \textbf{\methodname{}} \\
& (OOD) & (Failure) & \\
\midrule
\texttt{CoffeeSetupMug}       & 85.1 & 87.0 & \textbf{99.4} \\
\texttt{PnPCounterToCab}      & 43.1 & 56.2  & \textbf{73.7} \\
\texttt{TurnOffMicrowave}     & 54.9  & 50.0  & \textbf{74.5} \\
\texttt{TurnOffSinkFaucet}    & 11.3 & 38.8 & \textbf{45.0} \\
\texttt{OpenDoorDoubleHinge}  & 42.8  & 63.3 & \textbf{78.4} \\
\bottomrule
\end{tabular}
\vspace{2mm}
\caption{\textbf{Human Intervention Overlap Accuracy (\%) for Different Tasks.} \methodname{} (using OOD + Failure combined) achieves better performance than using OOD prediction only or Failure prediction only.}
\label{tab: ablation-ood-failure}
\end{table}

\begin{figure}[h!]
    \centering
    \includegraphics[width=0.85\linewidth]{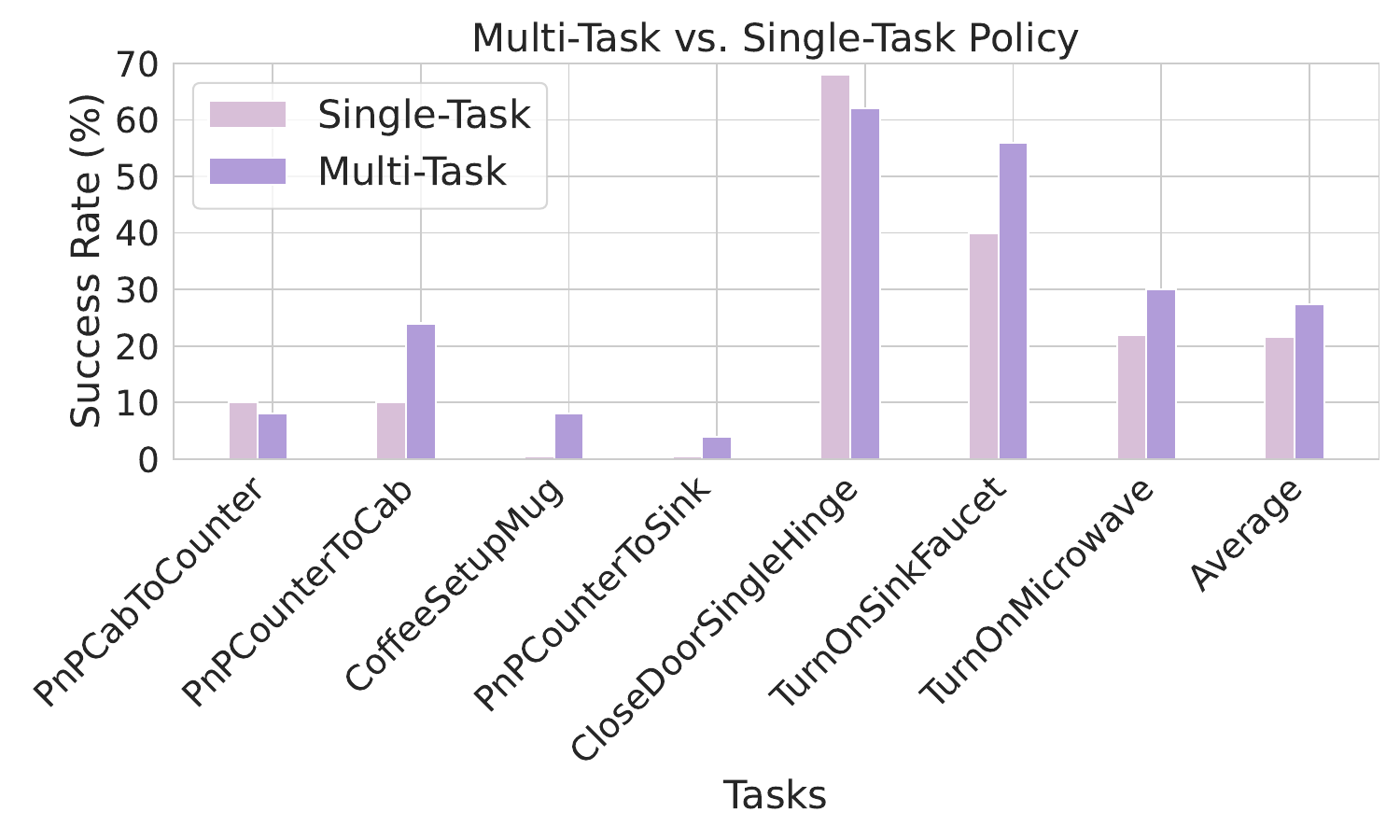}
    \caption{\textbf{Multi-task Policy vs. Single-task Policy Comparison.} We show the policy success rate (\%) for different tasks, comparing the multi-task policy trained on all tasks with single-task policies trained on individual tasks. \methodname{} 's multi-task policy gives overall better performance. Note that for \texttt{CoffeeSetupMug} and \texttt{PnPCounterToSink}, the single-task policy gives $0$\% success rate.}
    \label{fig:multi-single-policy}
\end{figure}
\clearpage
\clearpage
\subsection{Implementation Details}
\label{appendix: impl-details}

We present the details of hyperparameters for training the world model, policy, and error predictors in the following table:
\begin{itemize}
    \item World Model Architecture: Table \ref{table:hp-world-model};
    \item Policy Architecture: Table \ref{table:hp-policy};
    \item Training: Table \ref{table:training};
    \item Failure Prediction: Table \ref{table:failure-detection};
    \item OOD Prediction: Table \ref{table:ood-detection}.
\end{itemize}

\begin{table*}
\vspace{15pt}
\centering
\caption{Hyperparameters: World Model}
\begin{tabular}{c c c}
\hline
\textbf{Group} & \textbf{hyperparameter} & \textbf{Value}\\
\hline
& \\
\textbf{UNet Encoder} & image output activation & Sigmoid\\
& image observations & [workspace, wrist] \\
& observation fusion method & concat\\
& input channels & 3\\
& output channels & 3\\
& latent channels & 4\\
& block output channels & [32, 64]\\
& layers per block & 1\\
& activation function & SiLU\\
& normalization number of groups & 32\\
& \\
\textbf{Dynamics Model} & history length & 10\\
& number of futures sampled & 20\\
& \\
\textbf{GMM Prior} & latent dimension & 16\\
& learnable & True\\
& number of GMM nodes & 10\\
& \\
\textbf{Transformer Architecture} & context length & 20\\
\textbf{for VAE Encoder, Decoder, Prior} & embed dimension & 512\\
 & number of layers & 6\\
 & number of heads & 8\\
 & embedding dropout & 0.1\\
 & attention dropout & 0.1\\
 & block output dropout & 0.1\\
 & sinusoidal embedding & False\\
 & activation & GeLU\\
 & causal & False\\
& \\
\hline
\end{tabular}
\label{table:hp-world-model}
\end{table*}

\begin{table*}
\vspace{15pt}
\centering
\caption{Hyperparameters: Policy}
\begin{tabular}{c c c}
\hline
\textbf{Group} & \textbf{Hyperparameter} & \textbf{Value}\\
\hline
& \\
\textbf{Transformer Policy} & context length & 20\\
 & embed dimension & 512\\
 & mumber of layers & 6\\
 & number of heads & 8\\
 & embedding dropout & 0.1\\
 & attention dropout & 0.1\\
 & block output dropout & 0.1\\
 & sinusoidal embedding & False\\
 & activation & GeLU\\
 & causal & False\\
 & \\
\textbf{GMM Head} & number of modes & 5\\
 & min std & 0.005\\
 & std activation & Softplus\\
 & low noise eval & True\\
 & \\ 
\textbf{Image Encoder} & feature dimension & 64\\
  & backbone class & ResNet18ConvFiLM\\
  & backbone pretrained & False\\
  & backbone input coord conv & False\\
  & pool class & SpatialSoftmax\\
  & pool number of keypoints & 32\\
  & pool learnable temperature & False\\
  & pool temperature & 1.0\\
  & pool noise std & 0.0\\
  & \\
\textbf{Image Augmentation} & class & CropRandomizer\\
  & crop height & 116\\
  & crop width & 116\\
  & number of crops & 1\\
  & positional encoding & False\\
& \\
\hline
\end{tabular}
\label{table:hp-policy}
\end{table*}

\begin{table*}
\vspace{15pt}
\centering
\caption{Hyperparameters: Training}
\begin{tabular}{c c c}
\hline
\textbf{Group} & \textbf{Hyperparameter} & \textbf{Value}\\
\hline
& \\
\textbf{World Model} & optimizer type & Adam\\
 & initial learning rate & 0.0001\\
  & learning rate decay factor & 0.1\\
  & epoch schedule & []\\
  & scheduler type & Constant\\
& \\
\textbf{Policy} & optimizer type & AdamW\\
 & initial learning rate & 0.0001\\
 & learning rate decay factor & 1.0\\
 & epoch schedule & [100]\\
 & scheduler type & Constant with warmup\\
& \\

\hline
\end{tabular}
\label{table:training}
\end{table*}

\begin{table*}
\vspace{15pt}
\centering
\caption{Hyperparameters: Failure Prediction}
\begin{tabular}{c c c}
\hline
\textbf{Group} & \textbf{Hyperparameter} & \textbf{Value}\\
\hline

& \\
\textbf{Transformer Architecture} & context length & 10\\
 & embed dimension & 512\\
 & number of layers & 6\\
 & number of heads & 8\\
 & embedding dropout & 0.1\\
 & attention dropout & 0.1\\
 & block output dropout & 0.1\\
 & sinusoidal embedding & False\\
 & activation & GeLU\\
 & causal & False\\
 & \\

\textbf{Failure Predictor} & predict on future & True\\
& threshold count & 2\\
& evaluation index & [7,8,9]\\
& use probability & False\\
 & \\
\hline
\end{tabular}
\label{table:failure-detection}
\end{table*}

\begin{table*}
\vspace{15pt}
\centering
\caption{Hyperparameters: OOD Prediction}
\begin{tabular}{c c c}
\hline
\textbf{Group} & \textbf{Hyperparameter} & \textbf{Value}\\
\hline
 & \\
\textbf{OOD Predictor} & number of future steps & 20\\
& predict on future & True\\
& distance metric & k-means\\
& train k-means & False\\
& percentile & see Section \ref{method: adaptive} \\
 & \\
\hline
\end{tabular}
\label{table:ood-detection}
\end{table*}

\clearpage
\clearpage

\subsection{Qualitative Analysis and Discussion}
\label{appendix: qual}
Can \methodname{} catch important errors and give meaningful intervention timings? We conduct a qualitative analysis of where the error predictors predict error during robot execution using \textit{active human monitoring}. Specifically, we have a human operator who fully supervises the robot's policy execution and can intervene whenever an unsafe state is observed. We then apply the learned error predictors to the collected trajectories and compare failure states identified by the error predictors with those intervened by the human operator. This shows how the error predicted aligns with humans' risk assessment.

We show examples of the human intervention region and errors predicted by \methodname{} and the baselines in Figure \ref{fig:qual_ours_fail} to \ref{fig:qual_thrifty}. We use the same trajectory for all methods. The three instances humans gave interventions are: 1) the robot is not aiming at the object it is grasping; 2) the robot pauses at grasping; 3) the arm is too stiff and does not bend down. We visualize if the errors given by \methodname{} and the baselines are aligned with the intervention humans gave.

The \textcolor{teal}{green} area indicates times of actual human intervention; the \textcolor{blue}{blue} plot indicates the error value predicted by the different methods. We also show the video frame at the corresponding timings to illustrate the failure modes captured (or missed) by the different methods.

Note that \methodname{}, PATO, and ThriftyDAgger each have two separate components, so we visualize each individually. PATO and ThriftyDAgger share the same ensemble uncertainty component, so we visualize one of them.

As shown in Figure \ref{fig:qual_ours_fail} and \ref{fig:qual_ours_ood}, our method can capture a similar error pattern to that the human judges. MoMaRT (Figure \ref{fig:qual_momart}) is prone to predicting false positives due to pixel changes. PATO (VAE part, Figure \ref{fig:qual_pato_vae}) often predicts false negatives; PATO (ensemble part, Figure \ref{fig:qual_pato_ens}) often has high oscillations and extreme values. ThriftyDAgger's (Figure \ref{fig:qual_thrifty}) risk Q function is data-intensive to train and often has generalization errors; it cannot accurately reflect the task progress and local failure modes.

\begin{figure}[h!]
    \centering
    \includegraphics[width=0.9\linewidth]{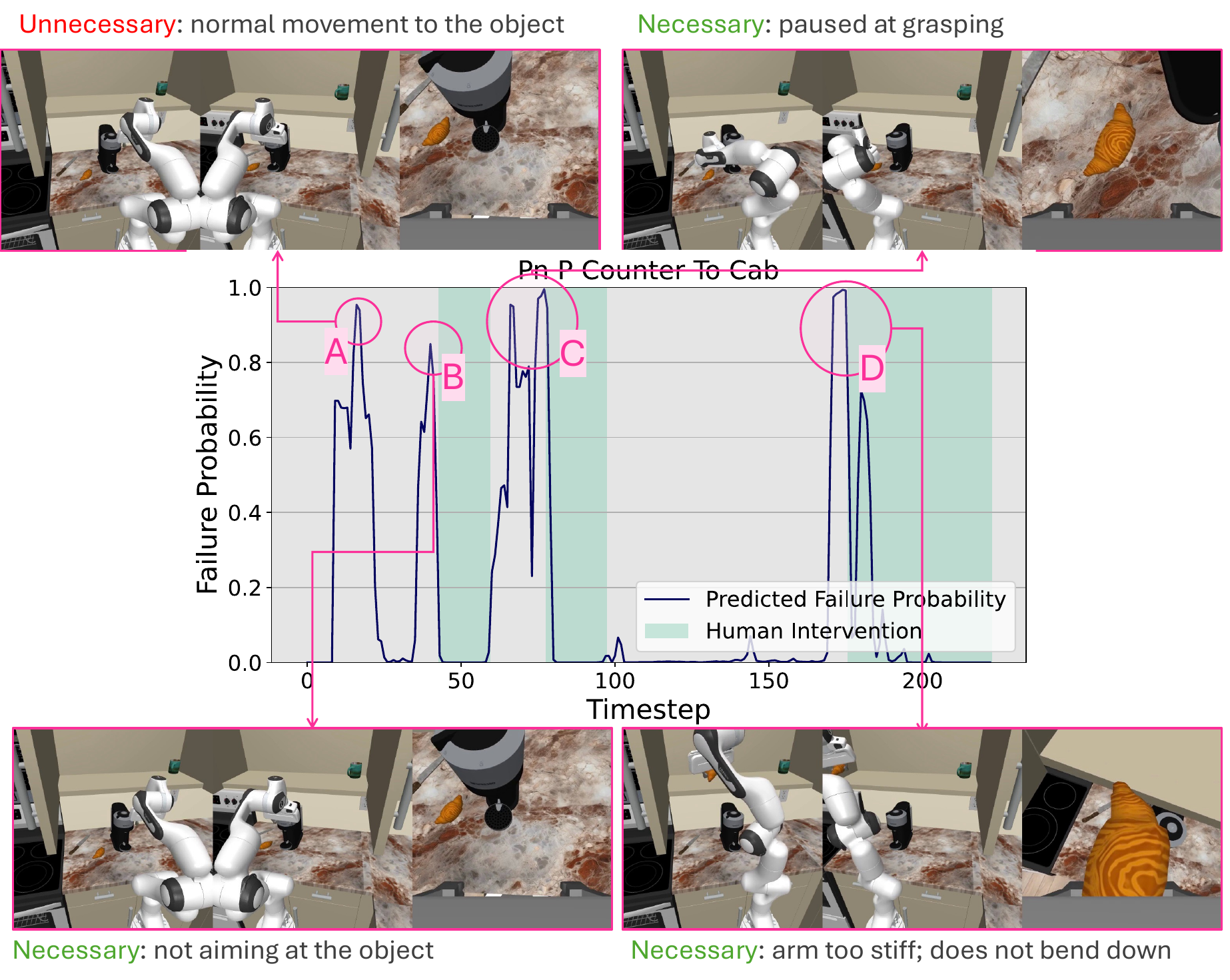}
    \caption{\textbf{\methodname{}: Failure Only.} The y-axis shows the predicted failure probability from the failure classifier, where a higher value indicates a greater likelihood of failure.}
    \label{fig:qual_ours_fail}
\end{figure}

\begin{figure}[h!]
    \centering
    \includegraphics[width=0.9\linewidth]{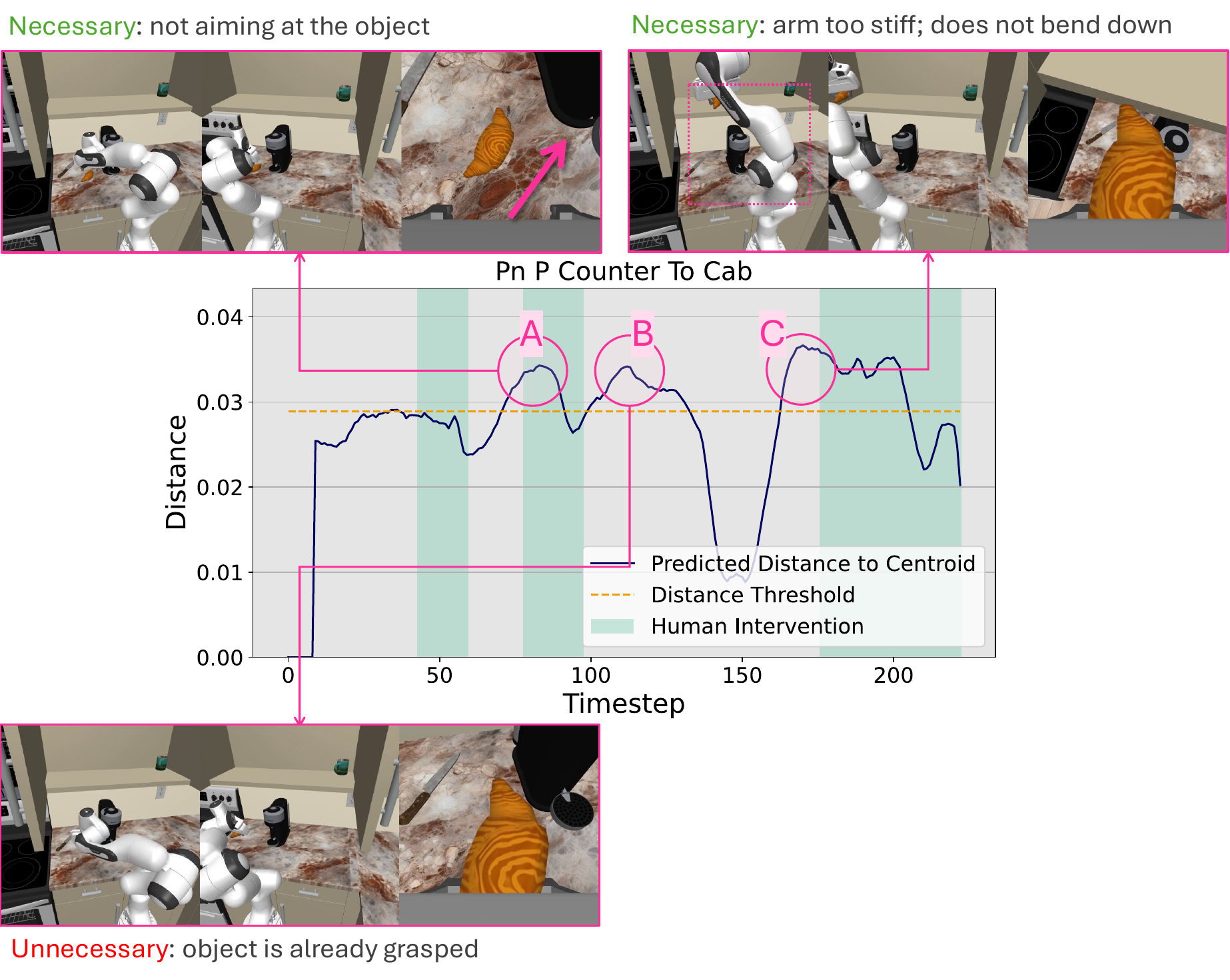}
    \caption{\textbf{\methodname{}: OOD Only.} The y-axis represents the predicted distance to the centroid. It is predicted as OOD if this distance exceeds a threshold (indicated by the yellow dotted line).}
    \label{fig:qual_ours_ood}
\end{figure}

\begin{figure}[h!]
    \centering
    \includegraphics[width=0.9\linewidth]{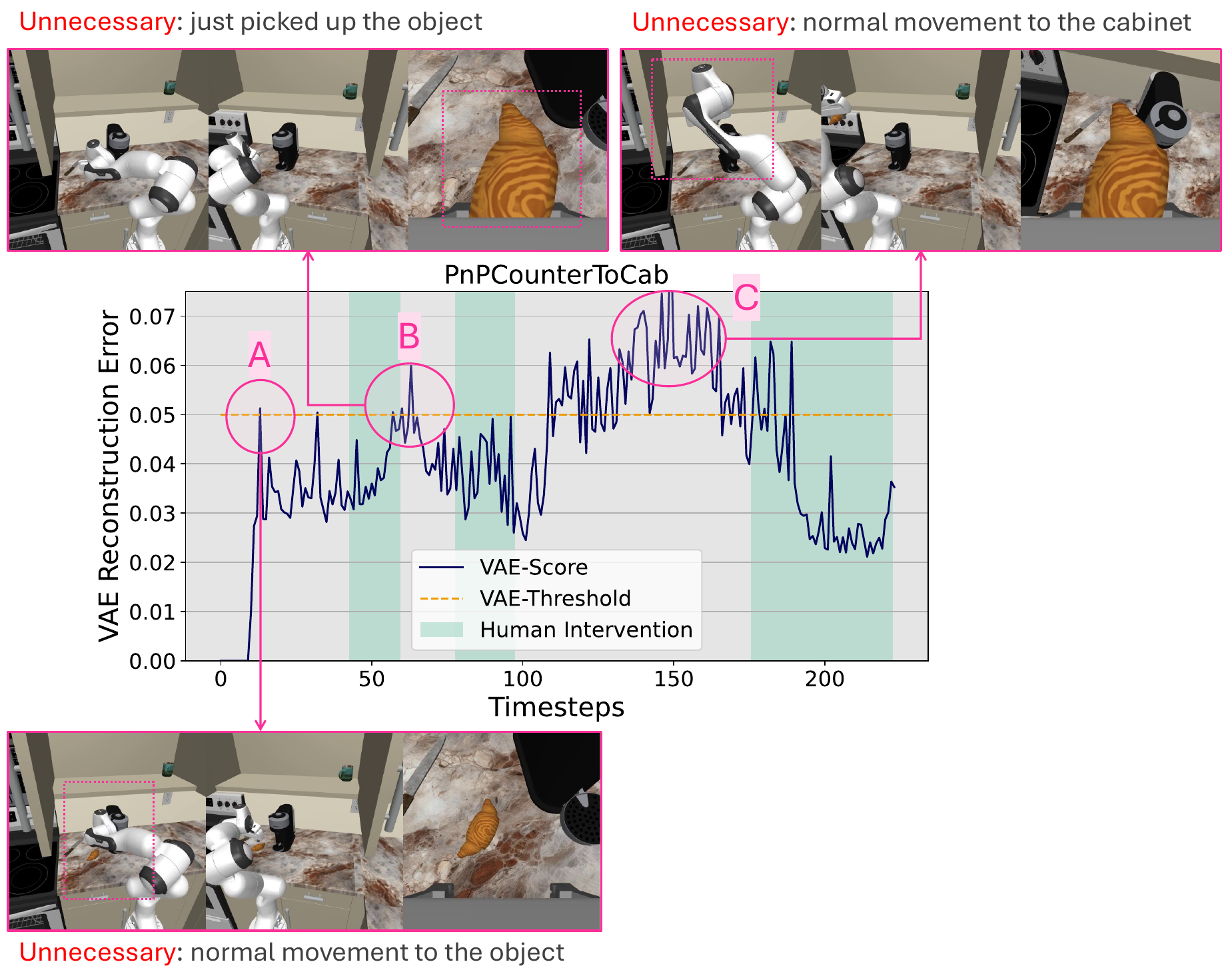}
    \caption{\textbf{MoMaRT.} The y-axis represents the VAE reconstruction error (MSE loss) of the current image observation. It is predicted as an error if this loss exceeds a threshold (indicated by the yellow dotted line).}
    \label{fig:qual_momart}
\end{figure}

\begin{figure}[h!]
    \centering
    \includegraphics[width=0.9\linewidth, trim=0 60 0 120, clip]{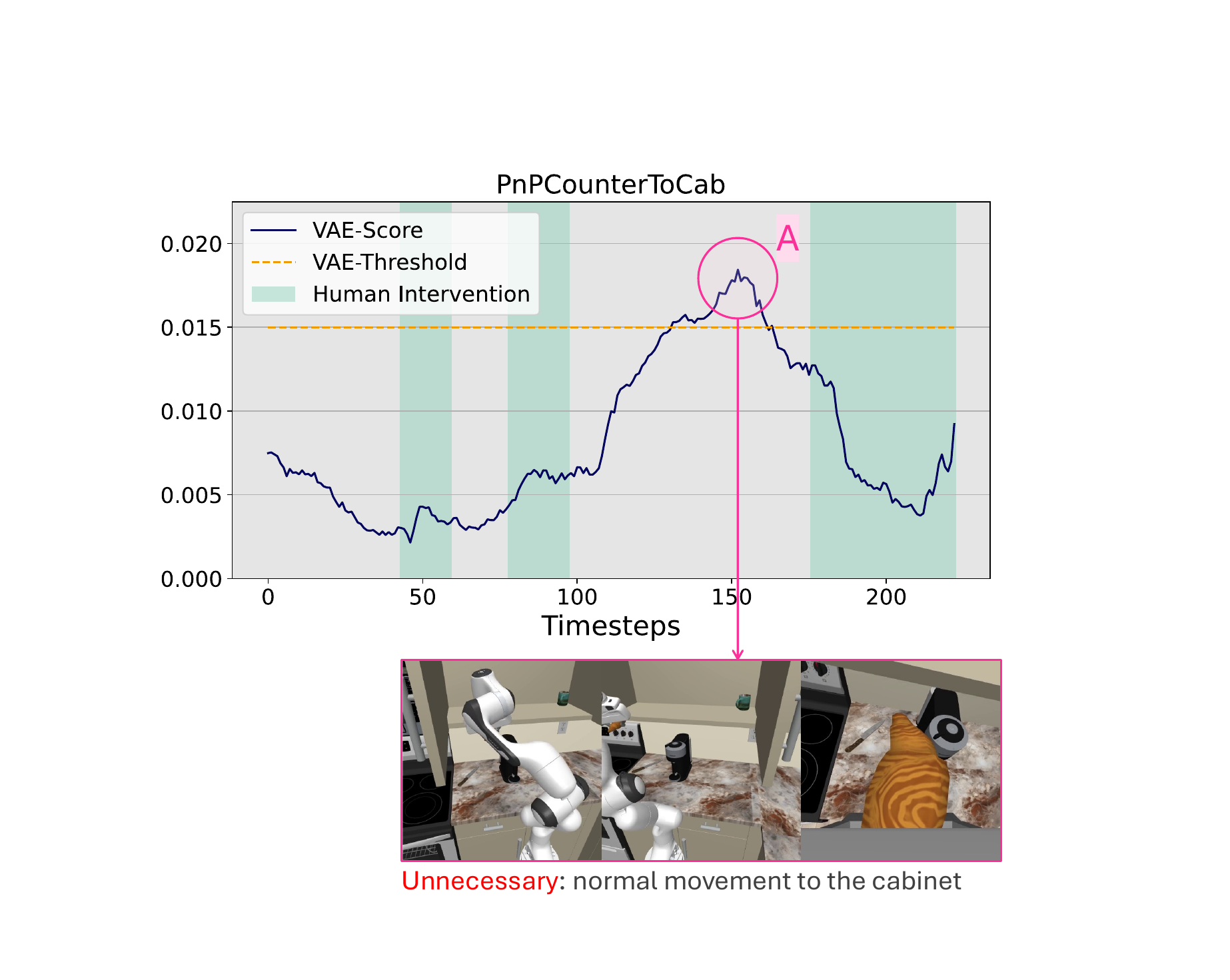}
    \caption{\textbf{PATO: VAE Only.} The y-axis represents the variance of the VAE reconstruction of the future image. It is predicted as an error if this value exceeds a threshold (indicated by the yellow dotted line).}
    \label{fig:qual_pato_vae}
\end{figure}

\begin{figure}[h!]
    \centering
    \includegraphics[width=0.9\linewidth]{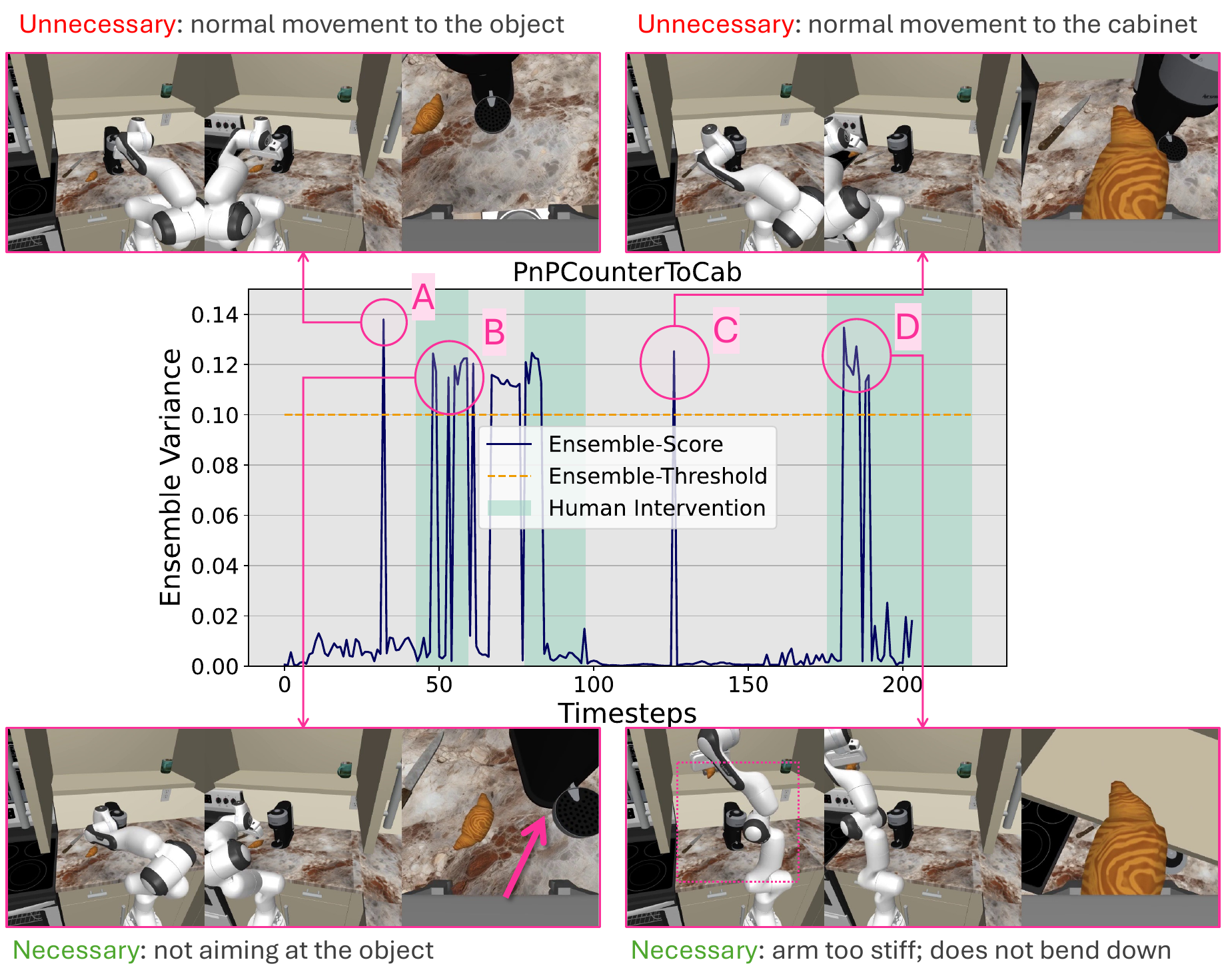}
    \caption{\textbf{PATO: Ensemble Only.} The y-axis represents the variance of the ensemble policies. It is predicted as an error if this value exceeds a threshold (indicated by the yellow dotted line).}
    \label{fig:qual_pato_ens}
\end{figure}

\begin{figure}[h!]
    \centering
    \includegraphics[width=\linewidth, trim=0 280 0 0, clip]{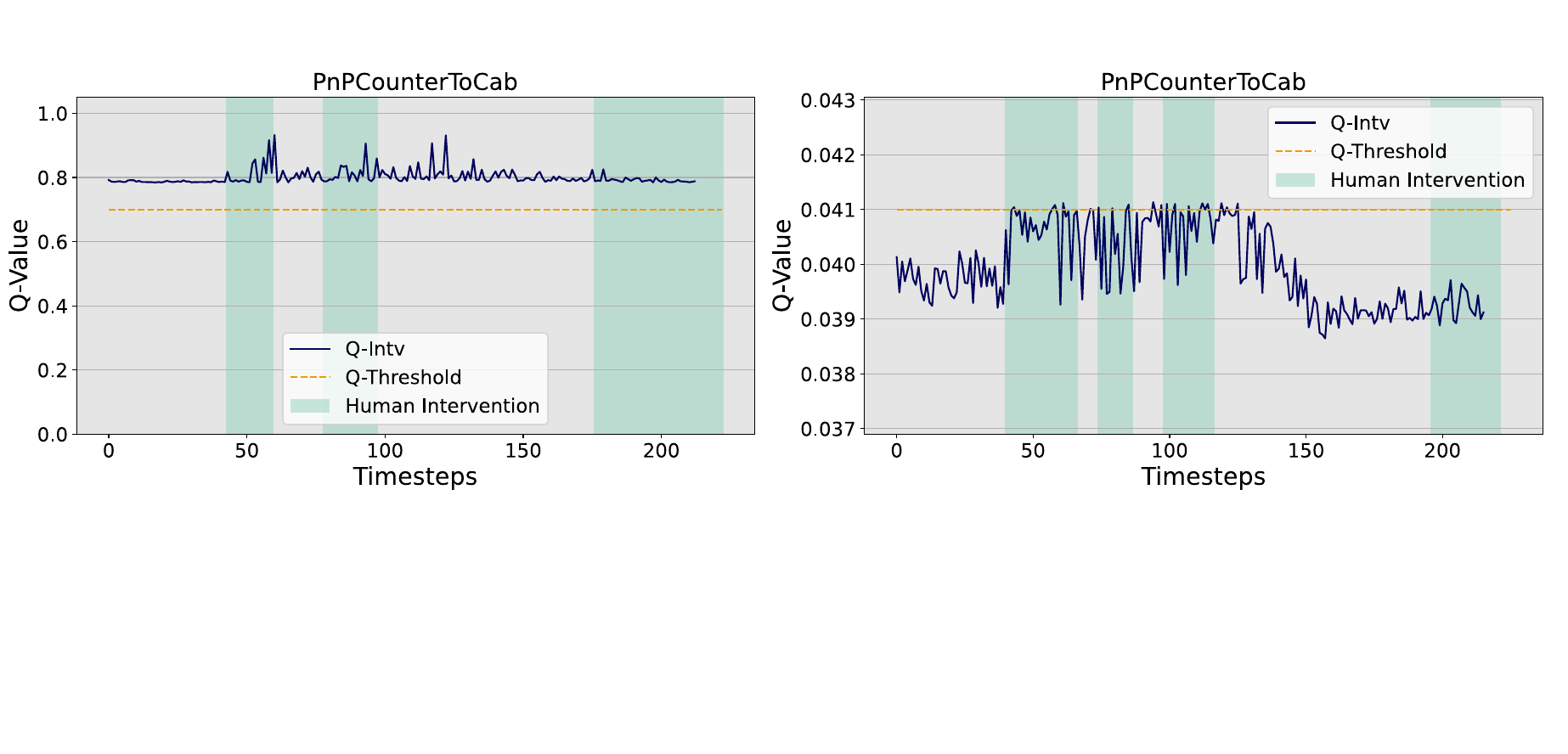}
    \caption{\textbf{ThriftyDAgger: Risk Q-Function.} The y-axis represents the q-value of the current state-action pair, where a higher q-value means less risk of failure. The video frame visualization is omitted here since no clear error modes have been predicted.}
    \label{fig:qual_thrifty}
\end{figure}

\clearpage

\subsection{Additional Details on Tasks}
\label{appendix: tasks}

\subsubsection{Real robot experiments}

\textbf{Dataset.} We use the Mutex \cite{shah2023mutex} dataset for real robot experiments. The Mutex dataset is a diverse multi-task dataset containing $50$ different tasks from $8$ task suites (where one task suite is a distinct set of objects and receptacles), specified with natural language. More details on the Mutex dataset can be found in the original Mutex \cite{shah2023mutex} paper. We use the original $50$ tasks for training the visual world model; for human workload consideration, we sample $10$ of the tasks (among $5$ task suites) for multi-task policy learning and runtime monitoring at deployment. The list of $10$ tasks for policy learning and runtime monitoring experiments is shown in Figure \ref{fig:task_vis_mutex}.

\begin{figure}[h!]
    \centering
    \includegraphics[width=\linewidth]{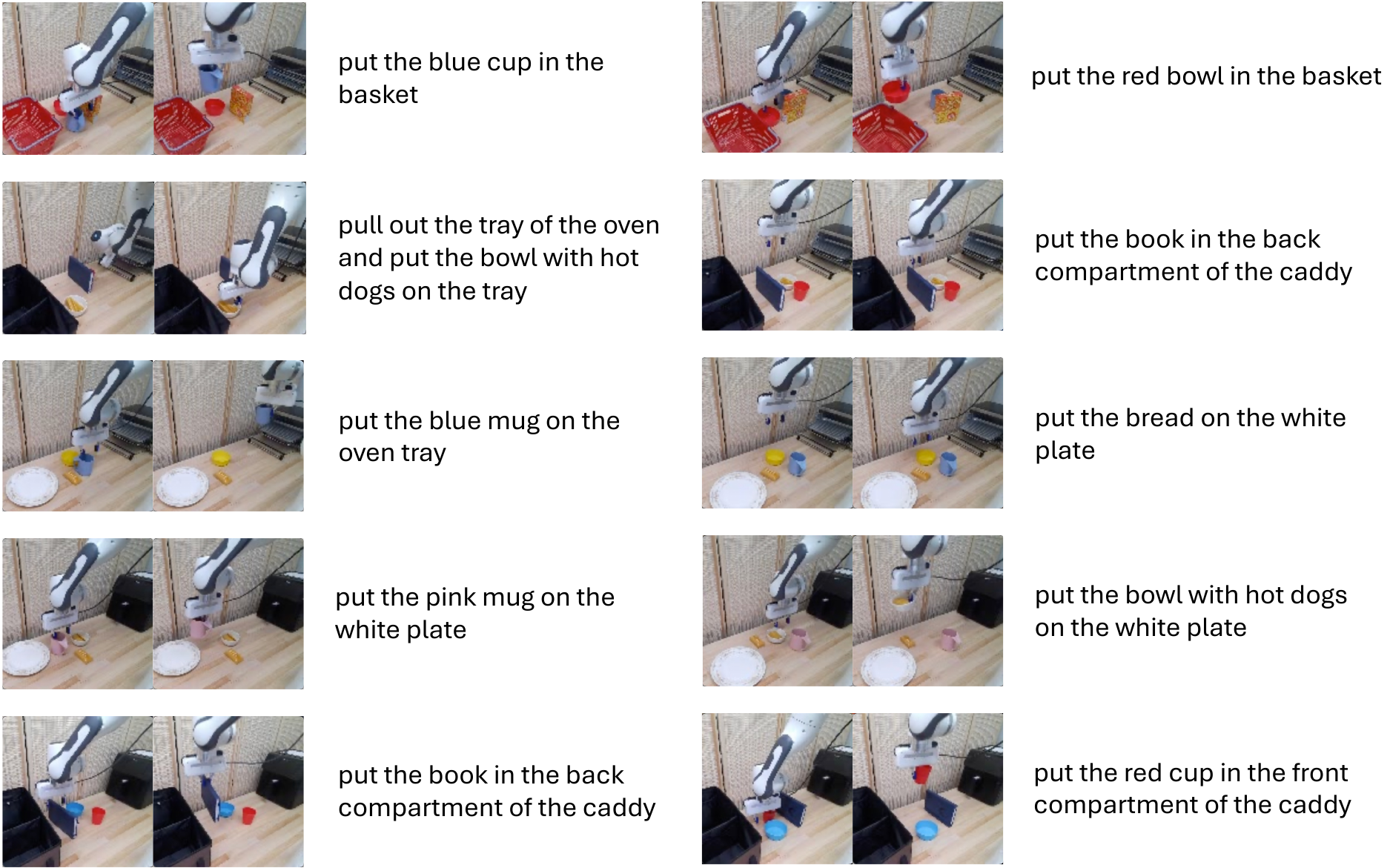}
    \caption{\textbf{Mutex Real World Tasks.} We use $10$ tasks covering $5$ task suites in the Mutex dataset for policy learning and runtime monitoring.}
    \label{fig:task_vis_mutex}
\end{figure}

\textbf{Training.} To train the visual world model, we use $30$ demonstrations for each of the $50$ tasks. To train the multi-task policy, $30$ demonstrations per task are used to bootstrap the initial BC policy; then, for each of the human-in-the-loop deployment round for runtime monitoring (see Evaluation Setting in Section \ref{exp:evaluation_setting}), $60$ robot rollouts per task are collected for each round. 

For the initial BC policy, we train the BC transformer policy for $2000$ epochs. For each subsequent round, we finetune the initial BC policy on the newly aggregated data for another $800$ epochs. We use this finetuned BC policy checkpoint to perform runtime monitoring experiments. 

\textbf{Evaluation.} We use $1$ seed and the checkpoint at a fixed epoch (epoch $2000$ for the initial policy, epoch $800$ for the finetuned per-round policy) to evaluate the real robot policy. We evaluate all $10$ tasks, conduct $20$ trials per task for $200$ trials, and report the average success rate across tasks. We report the per-task autonomous policy performance in Figure \ref{fig:Mutex_task_round_scores} and per-task combined policy performance in Figure \ref{fig:Mutex_task_round_scores_combined} for reference.

\subsubsection{Simulation}

\textbf{Dataset.} We use RoboCasa \cite{robocasa2024} as our simulation environments, which contains a diverse range of tasks  \textit{We note one important difference in the environment definition in RoboCasa: the environments are defined by a group of tasks that has similar task semantics, rather than by one single task of a fixed set of scenes and objects.} For example, The \texttt{CoffeeServeMug} environment can contain a combination of diverse backgrounds, layouts, coffee machine types, and coffee mug types, making it a challenging environment for generalization. As noted in Section \ref{exp:evaluation_envs}, We use all $20$ suites to train the visual world model and $12$ of them for policy learning and runtime monitoring. The list of $10$ tasks for policy learning and runtime monitoring experiments is shown in Figure \ref{fig:task_vis_robocasa}.

\begin{figure}[h!]
    \centering
    \includegraphics[width=\linewidth]{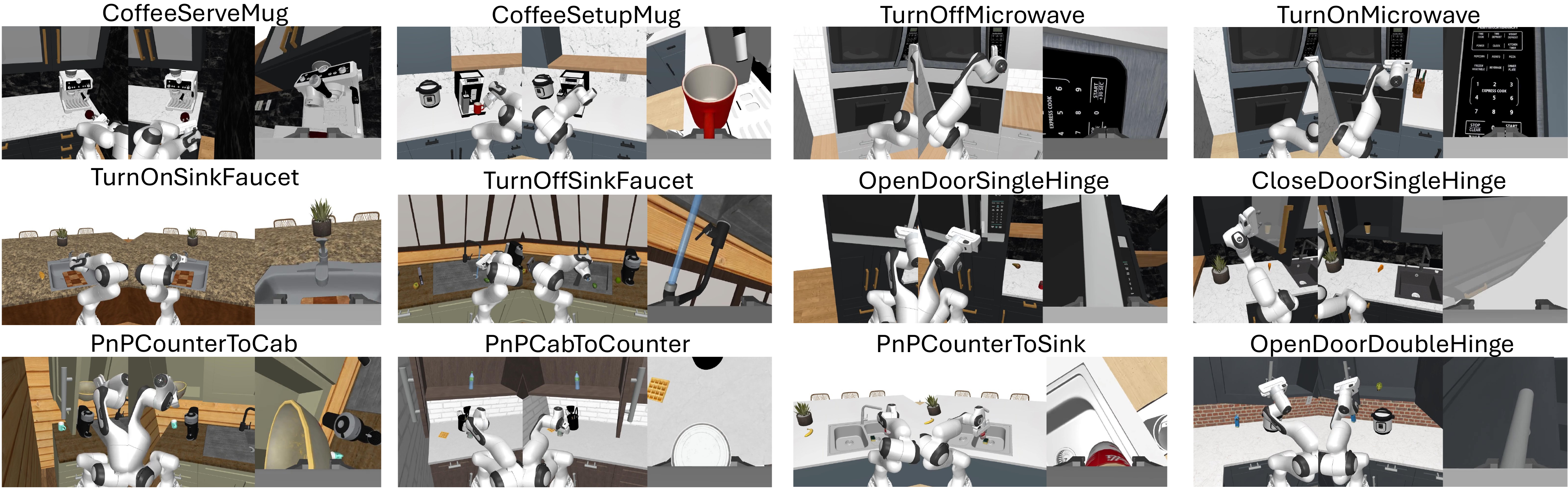}
    \caption{\textbf{RoboCasa Simulation Tasks.} We use $12$ tasks in the RoboCasa benchmark for policy learning and runtime monitoring.}
    \label{fig:task_vis_robocasa}
\end{figure}

\textbf{Training.} To train the visual world model, we use $50$ demonstrations for each of the $20$ tasks suites. To train the multi-task policy, $50$ demonstrations per task suite are used to bootstrap the initial BC policy; then, for each of the human-in-the-loop deployment round for runtime monitoring (see Evaluation Setting in Section \ref{exp:evaluation_setting}), $100$ robot rollouts per task are collected for each round. 

For the initial BC policy, we train the BC transformer policy for $1000$ epochs. For each subsequent round, we finetune the initial BC policy on the newly aggregated data for another $400$ epochs. We use this finetuned BC policy checkpoint to perform runtime monitoring experiments. 

\textbf{Evaluation. } We use $2$ seeds and checkpoints at a fixed epoch (epoch $1000$ for the initial policy, epoch $400$ for the finetuned per-round policy) to evaluate the simulation policy. We evaluate all $12$ tasks, conduct $50$ trials per task for a total of $600$ trials, and report the average success rate across tasks. We report the per-task autonomous policy performance in Figure \ref{fig:Robocasa_task_round_scores} and per-task combined policy performance in Figure \ref{fig:Robocasa_task_round_scores_combined} for reference.

\clearpage

\begin{figure}[t]
    \centering
    \includegraphics[width=\linewidth]{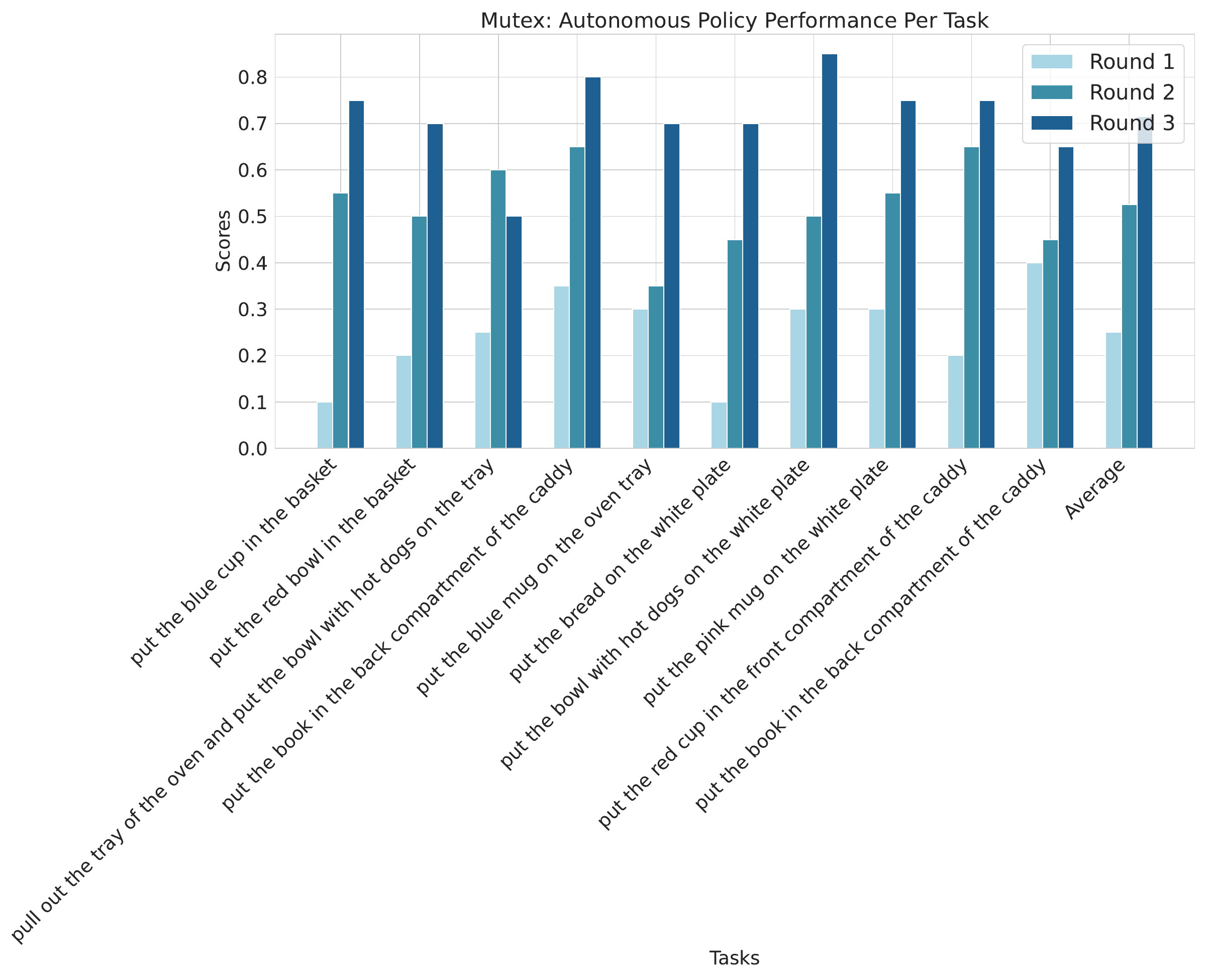}
    \caption{\textbf{Mutex: Autonomous Policy Performance Per Task.} The autonomous policy success rate improves over three rounds of deployment for the Mutex tasks.}
    \label{fig:Mutex_task_round_scores}
\end{figure}

\begin{figure}[t]
    \centering
    \includegraphics[width=\linewidth]{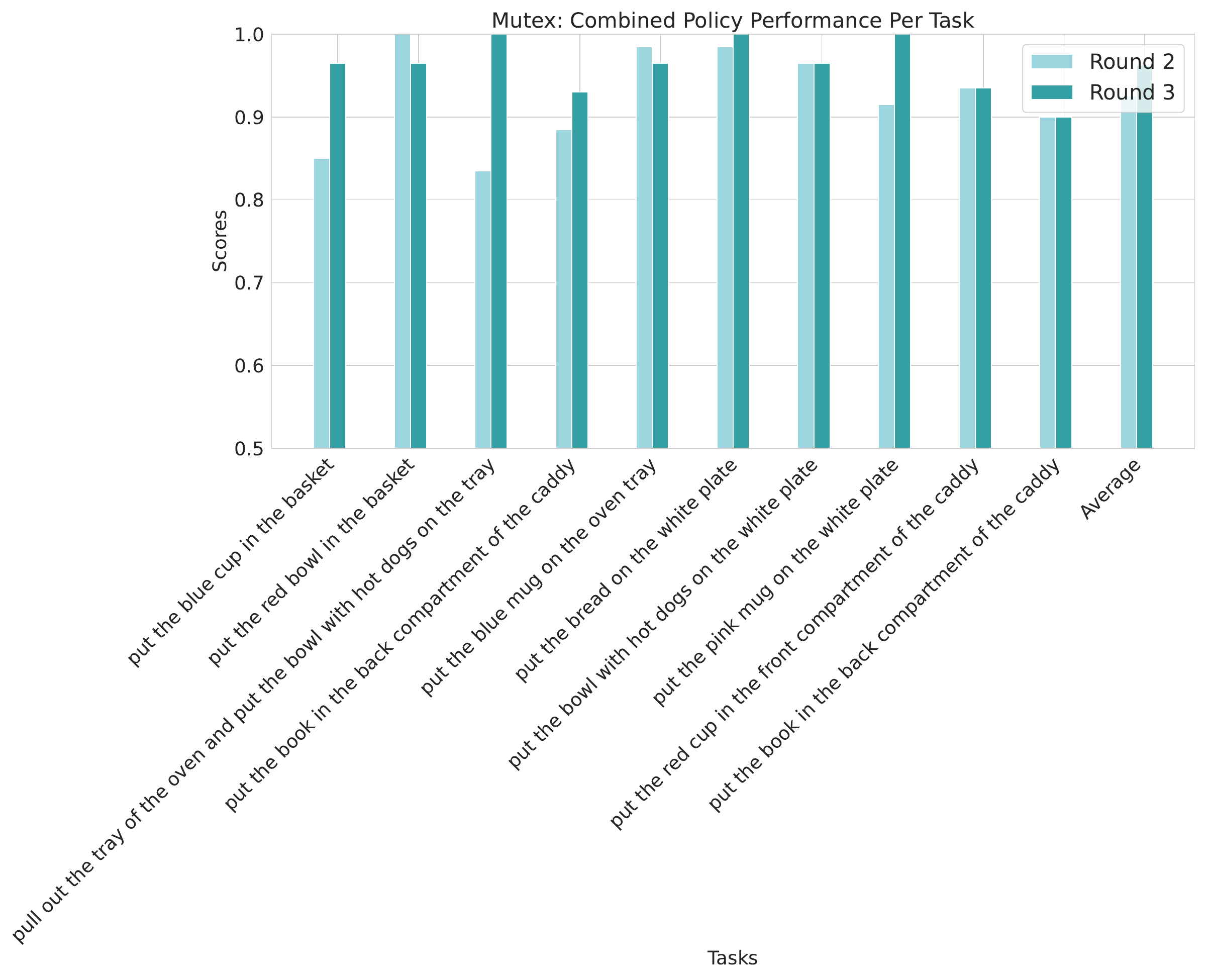}
    \caption{\textbf{Mutex: Combined Policy Performance Per Task.} The combined policy success rate improves over three rounds of deployment for the Mutex tasks.}
    \label{fig:Mutex_task_round_scores_combined}
\end{figure}

\begin{figure}[t]
    \centering
    \includegraphics[width=\linewidth]{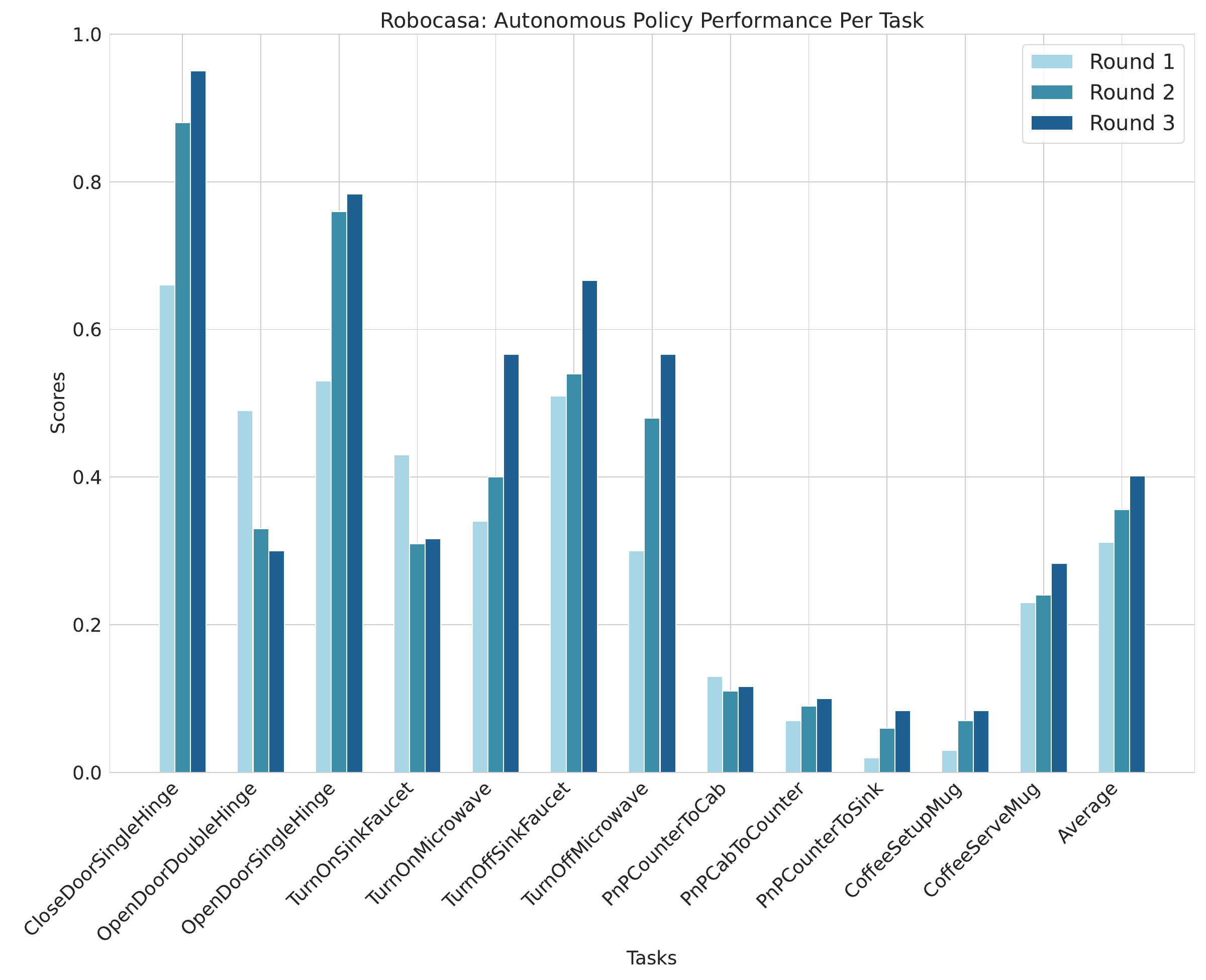}
    \caption{\textbf{RoboCasa: Autonomous Policy Performance Per Task.} The autonomous policy success rate improves over three rounds of deployment for most of the RoboCasa tasks.}
    \label{fig:Robocasa_task_round_scores}
\end{figure}

\begin{figure}[t]
    \centering
    \includegraphics[width=\linewidth]{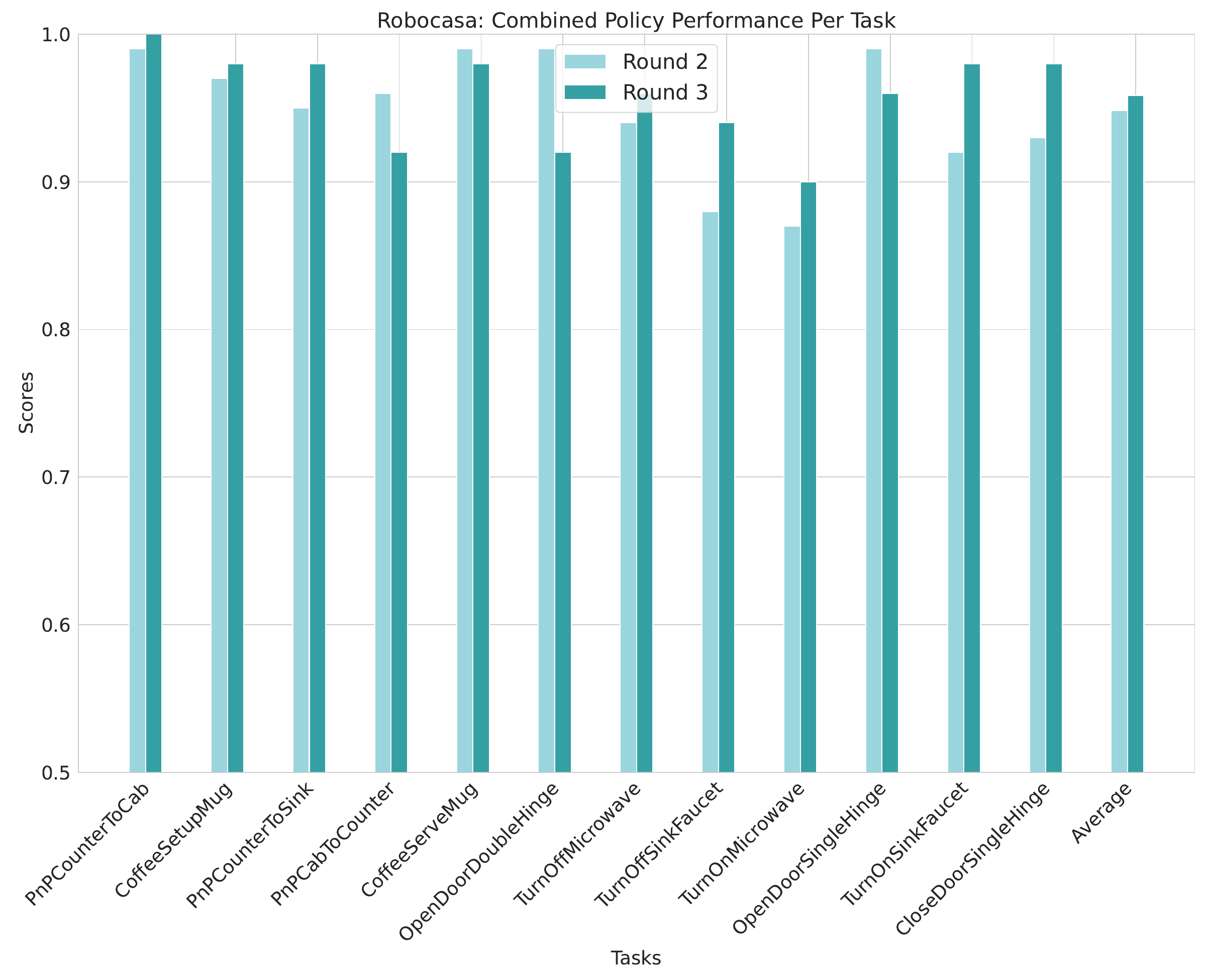}
    \caption{\textbf{RoboCasa: Combined Policy Performance Per Task.} The combined policy success rate improves over three rounds of deployment for most of the RoboCasa tasks.}
    \label{fig:Robocasa_task_round_scores_combined}
\end{figure}





\clearpage

\end{document}